\documentclass{article}
\usepackage{kr}

\usepackage{times}
\usepackage{soul}
\usepackage{url}
\usepackage[hidelinks]{hyperref}
\usepackage[T1]{fontenc}
\usepackage[utf8]{inputenc}
\usepackage[small]{caption}
\usepackage{graphicx}
\usepackage{amsmath,amssymb}
\usepackage{booktabs}
\usepackage{makecell}
\usepackage{microtype}
\usepackage{placeins}
\usepackage{rotating}
\usepackage{titlesec}
\titlespacing*{\textbf}{0pt}{1.0ex plus 0.5ex minus 0.2ex}{1em}

\newcommand{\Ab}{\mathsf{Ab}}

\newcommand{\TODO}[1]{\textsuperscript{$\dagger$}}

\usepackage{listings}
\usepackage{xcolor}
\usepackage{enumitem}

\usepackage{dblfloatfix}

\definecolor{promptbg}{RGB}{248,248,248}
\definecolor{promptframe}{RGB}{180,180,180}
\lstdefinestyle{prompt}{
  basicstyle=\ttfamily\scriptsize,
  backgroundcolor=\color{promptbg},
  frame=single,
  rulecolor=\color{promptframe},
  breaklines=true,
  breakatwhitespace=false,
  columns=fullflexible,
  keepspaces=true,
  tabsize=2,
  showstringspaces=false,
  xleftmargin=4pt,
  xrightmargin=4pt,
  aboveskip=8pt,
  belowskip=8pt,
  extendedchars=true,
  inputencoding=utf8,
  literate=
    {α}{{\ensuremath{\alpha}}}1
    {₁}{{\ensuremath{_1}}}1
    {₂}{{\ensuremath{_2}}}1
    {→}{{\ensuremath{\to}}}1
    {↔}{{\ensuremath{\leftrightarrow}}}1
    {¬}{{\ensuremath{\lnot}}}1
    {∧}{{\ensuremath{\land}}}1
    {∃}{{\ensuremath{\exists}}}1
    {∀}{{\ensuremath{\forall}}}1
    {Σ}{{\ensuremath{\Sigma}}}1,
}

\begin{document}

\title{\textsc{ABD}: Default--Exception Abduction in Finite First-Order Worlds}
\author{Serafim Batzoglou\footnote{Contact: serafim.batzoglou@gmail.com}}
\maketitle

\begin{abstract}
Abduction in knowledge representation is often framed as ``explaining away'' inconsistencies between
a background theory and observations by hypothesizing missing facts or exceptions.
Despite decades of KR work on abduction, there are few modern benchmarks with exact, solver-checkable semantics for multi-world first-order exception synthesis.
We introduce \textsc{ABD}, a solver-checkable benchmark for default--exception abduction over small finite relational worlds, and use it 
here for a diagnostic evaluation of frontier language models.
Each instance provides (a) a set of finite structures with observed facts, and (b) a fixed
default-like first-order theory that may be violated by those observations.
A model must output a first-order \emph{abnormality rule} $\alpha(x)$ whose substitution for $\Ab$
restores satisfiability while keeping exceptions sparse.

We formalize three observation regimes with distinct completion semantics.
\textbf{ABD-Full} assumes closed-world observation.
\textbf{ABD-Partial} allows unknown atoms under \emph{existential} completion: $\alpha$ is valid 
if \emph{some} completion makes the repaired theory satisfiable, with cost optimized in the best case.
\textbf{ABD-Skeptical} uses \emph{universal} completion: $\alpha$ is valid only if the repaired theory 
is satisfiable under \emph{every} completion, with cost measured in the worst case.

Because domains are finite, validity and costs are computed via SMT (Z3), enabling exact verification 
and controlled difficulty.
In this paper, we use \textsc{ABD} to evaluate eleven frontier LLMs on 600 instances spanning all three scenarios and seven default theories.
Among the strongest high-validity models, prompt-set cost gaps of ${\sim}1$--$1.5$ extra 
exceptions per world remain, and valid-formula AST stays in the low teens, while GPT-5.4 is a low-gap, high-AST outlier with lower validity.
Holdout evaluation reveals distinct generalization profiles: in ABD-Full and ABD-Partial, the dominant failure is \emph{parsimony inflation}; in ABD-Skeptical, it is \emph{validity brittleness}---rules that work on prompt worlds often break on holdouts---while survivors show smaller gap inflation.
\end{abstract}

\section{Introduction}

Default reasoning is a cornerstone of knowledge representation: we often model domains with
rules that hold ``normally,'' while allowing rare exceptions.
In classical KR formalisms---default logic \cite{reiter1980default}, circumscription \cite{mccarthy1980circumscription}, and stable-model semantics \cite{gelfond1988stable}---exceptions are
not bugs but a feature: they encode domain irregularities without abandoning general rules.
From the perspective of \emph{abduction} \cite{poole1989explanation,eiter1995complexity}, exceptions are precisely what we infer when observations conflict
with what a default theory predicts.

Recent work shows that LLMs can emit first-order logical forms and support logic-theory induction under formal feedback \cite{PeiEtAl2025FoVer,GandarelaEtAl2025Expressivity}.
\emph{Can such models propose correct, compact exception rules that repair default theories across multiple
relational worlds?}
Answering this requires an evaluation setting that removes natural-language ambiguity, supports quantifiers,
and remains mechanically checkable.

We propose \textsc{ABD}, a solver-checkable benchmark for default--exception abduction on finite first-order worlds.
The benchmark is general in the sense that any system that outputs a first-order abnormality rule can be evaluated on it; the empirical focus here is a diagnostic study of frontier language models.
Each instance consists of several small relational structures (``worlds'') and a fixed first-order background
theory $\Theta$ that uses an abnormality predicate $\Ab$ to block defaults.
Rather than providing labeled targets, we provide only the theory and the observed facts; the model must output a definition of abnormality---a first-order formula $\alpha(x)$ such that substituting $\alpha$ for $\Ab$ repairs the theory across all prompt worlds.
Evaluation then combines validity with a parsimony objective: among valid hypotheses, we prefer rules that mark as few abnormal domain elements as possible.

The output $\alpha(x)$ is therefore best viewed as a shared repair policy for a default theory over multiple relational worlds.
The task is to restore consistency with minimal exceptional structure, not to recover a unique planted target formula.

\textbf{Three Observation Regimes.}
We study three scenarios that share the same interface but differ in how missing information is handled:
\textbf{ABD-Full} (closed-world), \textbf{ABD-Partial} (existential completion: valid if \emph{some} completion satisfies the repaired theory), and \textbf{ABD-Skeptical} (universal completion: valid only if \emph{every} completion satisfies the repaired theory).
Parsimony mirrors these quantifiers: best-case cost in ABD-Partial, worst-case in ABD-Skeptical.
These regimes are formalized in Section~\ref{sec:abduction}.

\textbf{Evaluation Beyond Validity.}
Binary validity quickly saturates for stronger models. However, validity is in principle trivial in 
this task: a model can simply label many (or all) elements as abnormal. We therefore emphasize \emph{cost-based} metrics---how 
close a model comes to a solver-computed lower bound on abnormality count---and \emph{size-conditioned} variants 
that group formulas by syntactic complexity (AST size) to discourage degenerate case-splitting.

\textbf{Main Empirical Findings.}
We evaluate eleven frontier models across all three regimes. Holdout evaluation reveals two distinct failure 
modes: \emph{parsimony inflation} in ABD-Full/Partial (gaps roughly double on fresh worlds) 
and \emph{validity brittleness} in ABD-Skeptical (rules that work on prompt worlds often break on holdouts, though 
survivors show smaller gap inflation). The models also separate into distinct performance profiles:
Opus-4.6, Gemini-3.1, DSR, and Grok-4.1-fast (Grok4.1f) form a high-validity cluster; GPT-5.4 attains the best gap metrics
but with very large formulas and weak holdout survival; Kimi-K2-thinking (Kimi-K2t) is comparatively compact and conditionally
robust but substantially less parsimonious. Details are in Section~\ref{sec:experiments}.

\textbf{Contributions.}
\begin{itemize}
  \item We formalize default--exception abduction on finite first-order worlds with solver-checkable
        semantics under three observation regimes: ABD-Full (closed-world, fully observed),
        ABD-Partial (existential completion under missing atoms), and ABD-Skeptical (universal completion with
        worst-case, robust exception costs).
  \item We introduce cost-based scoring for parsimony, including gap-to-lower-bound, reference-relative gap,
        and size-conditioned analyses by formula complexity.
  \item We describe a difficulty-controlled generator that constructs multi-world instances and
        eliminates shortcut hypotheses via a lightweight, counterexample-guided (CEGIS-like) procedure.
  \item We evaluate eleven frontier models---Opus-4.6, Grok4, Grok4.1f, Gemini-3.1, Gemini-3, GPT-5.2, GPT-5.4, DeepSeek-Reasoner (DSR), Kimi-K2t, GPT-4o, and Hermes4---and analyze validity, parsimony gaps, formula complexity, and holdout generalization.
\end{itemize}

\section{Related Work}\label{sec:related}

\textbf{Abduction, Defaults, and Non-Monotonic Reasoning.}
Abduction is often formalized as selecting hypotheses that, when added to background knowledge, entail observations while satisfying integrity constraints, typically with a preference for parsimonious explanations \cite{poole1989explanation,kakas1992abductive,eiter1995complexity}.
This connects directly to non-monotonic KR frameworks for defeasible rules and exceptions, including default logic and circumscription \cite{reiter1980default,mccarthy1980circumscription}, and stable-model/answer-set semantics that support explicit minimization principles (including ``minimize abnormalities'' encodings) \cite{gelfond1988stable,baral2003kr,brewka2011answer,gebser2012answer}.
Recent benchmark work has also tested LLMs directly on ASP solving, showing that full answer-set computation remains substantially harder than ASP entailment or answer-set verification \cite{RenEtAl2025ASPBench}.
Our benchmark adopts a solver-friendly abnormality-predicate encoding with finite grounding, but evaluates more than satisfiability: we quantify \emph{parsimony gaps} between a model's proposed exception rule and a solver-computed minimum-exception lower bound, and we test whether the learned exception policy remains valid on holdout worlds.

\textbf{Abductive Logic Programming and Diagnosis.}
Abductive logic programming (ALP) provides a general framework for explanation under integrity constraints and has been developed in both logic-programming and KR traditions \cite{kakas1992abductive,denecker2002abduction}.
Model-based diagnosis similarly treats faults/abnormalities as latent causes chosen to restore consistency, often with minimality criteria and hitting-set structure \cite{reiter1987theory,dekleer1987diagnosing,hamscher1992readings}.
Our setting differs in interface and emphasis: the hypothesis is an explicit \emph{first-order definition} of an abnormality predicate (not a set of abduced ground facts), and it must satisfy a \emph{multi-world} specification, which makes it possible to separate repair-by-case from compact exception structure that generalizes.

\textbf{Inductive Logic Programming and Relational Learning.}
ILP learns relational rules from examples and background knowledge, with classic systems and
formalisms for inducing Horn clauses and relational definitions \cite{muggleton1991inductive,muggleton1994inductive,quinlan1990learning,srinivasan2001aleph,cropper2020learning,raedt2008logical}.
Recent work has also studied LLMs as inductive theory learners under formal inference-engine feedback and graded target expressivity \cite{GandarelaEtAl2025Expressivity}, providing a complementary LLM-centered analysis of logic theory induction.
While our hypotheses are symbolic and relational, our tasks are not supervised
concept learning: each instance provides observations plus a default theory, and success depends
on proposing an exception rule that restores satisfiability while minimizing abnormalities.
This shifts the challenge from matching labels to balancing consistency and parsimony under an
explicit non-monotonic objective, with exact verification.
Synthesizing $\alpha(x)$ from multi-world specifications is also amenable to symbolic ILP and ASP-based learners such as Progol \cite{muggleton1995inverse} and ILASP \cite{law2014inductive}, which can search a hypothesis space with completeness guarantees relative to a language bias.
The benchmark is meaningful beyond LLMs, but our empirical scope here is LLM diagnosis rather than an exhaustive comparison across symbolic, ILP-, ASP-, or neuro-symbolic systems.

\textbf{Program Synthesis and Solver-Aided Search.}
Our tasks can be viewed as constrained synthesis: produce a symbolic artifact that satisfies a solver-checkable specification.
This aligns with solver-aided languages and synthesis paradigms such as Sketch, CEGIS, and syntax-guided synthesis \cite{solarlezama2006sketch,solarlezama2008thesis,alur2013sygus,torlak2014rosette}.
A key difference is that our specifications are \emph{multi-world} and explicitly \emph{costed}: hypotheses must simultaneously repair multiple structures, and evaluation tracks both validity and the gap between a model's abnormality cost and a solver-computed lower bound.

\textbf{Logic and Relational Reasoning Benchmarks.}
A broad range of benchmarks probe reasoning with rules, quantifiers, and structured knowledge, often through natural-language interfaces \cite{tafjord2021proofwriter,han2024folio,clark2020transformers}.
Such settings are valuable but can conflate logical competence with linguistic priors and make intermediate steps hard to verify. Recent work has also introduced InAbHyD, a synthetic benchmark for inductive and abductive reasoning with an Occam-style hypothesis metric over incomplete world models \cite{SunSaparov2025Occam}. Our benchmark instead presents extensional finite structures and uses solver-backed grounding to make correctness and costs exact, enabling diagnostics about exception sparsity, gap above a solver lower bound, and overfitting across holdout worlds.

\textbf{Mathematical Reasoning and Formal Proof Benchmarks.}
Natural-language math benchmarks such as GSM8K and MATH test multi-step reasoning but remain mediated by natural-language problem statements and outputs \cite{cobbe2021training,hendrycks2021measuring}.
Formal proof benchmarks (e.g., CoqGym, HOList, miniF2F, ProofNet, ProofGrid) evaluate mechanically verified outputs in proof assistants or formal proof languages \cite{yang2019learning,bansal2019holist,zheng2022minif2f,azerbayev2023proofnet,arkoudas-batzoglou-2025-stress}, and large-scale theorem-proving efforts use learning to guide deductive search \cite{polu2020generative}.
Our benchmark is complementary: rather than proving theorems, models must \emph{synthesize compact exception definitions} that reconcile defaults with observations under a minimum-abnormality objective.

\textbf{Neuro-Symbolic Learning and Large Language Models.}
Neuro-symbolic work explores combining learning with symbolic representations and constraint-based verification \cite{garcez2023neurosymbolic}, including approaches that use abduction as a learning signal \cite{dai2019abductivelearning}.
Relatedly, FoVer translates natural-language reasoning into first-order logic and uses automated logical verification to check logical correctness, illustrating the value of verification-backed evaluation for LLM reasoning claims \cite{PeiEtAl2025FoVer}.
Recent LLM analyses highlight that models can produce syntactically plausible symbolic outputs while relying on shortcuts or overly long solutions when unconstrained \cite{wei2022chain,dziri2023faith}.
Our evaluation emphasizes solver-verifiable semantics, parsimony-sensitive metrics (cost and gap), and holdout testing designed to expose hypotheses that repair observed worlds without capturing a stable general exception rule.

\textbf{Abductive Reasoning in Natural Language.}
Abduction has also been studied in NLP as a model of commonsense explanation and narrative understanding, including classic logical accounts of interpretation-as-abduction \cite{hobbs1993interpretation} and benchmark-style tasks for causal/abductive inference \cite{roemmele2011copa,mostafazadeh2016corpus,bhagavatula2020abductive}.
These tasks are valuable for modeling everyday explanation but are mediated by language and typically lack a fully explicit world model with exact cost verification.
By contrast, our benchmark isolates abductive structure in a formal finite-model setting where validity and costs are mechanically checkable and have a direct interpretation as exception cardinality.

\section{Problem Setup}
\label{sec:setup}

\textbf{Finite Relational Worlds.}
We fix a relational signature
\(
\Sigma=\{P,Q,R,S,=\}
\),
with unary predicates $P(\cdot),Q(\cdot)$, binary predicates $R(\cdot,\cdot),S(\cdot,\cdot)$, and equality.
A \emph{world} $W$ has a finite domain $D_W=\{a_1,\ldots,a_n\}$ ($|D_W|\in\{9{-}12\}$; see Table~\ref{tab:dataset_summary}) and (possibly partial) interpretations for
the predicates in $\Sigma$.
In the fully observed setting (ABD-Full), $W$ specifies complete interpretations
$P^W,Q^W\subseteq D_W$ and $R^W,S^W\subseteq D_W\times D_W$.
In the partial-observation settings (ABD-Partial and ABD-Skeptical), $W$ additionally designates a set of
\emph{unknown} ground atoms $\Omega_W$ whose truth values are not observed.
In our benchmark, $\Omega_W$ ranges over ground atoms of $R$ and $S$; $P$ and $Q$ are always fully observed.
In ABD, a world does \emph{not} include labels for a target concept.
This fixed signature and small-domain setting is deliberate: it keeps all three semantics groundable and SMT-checkable, so our claims are about this finite-world benchmark family rather than arbitrary first-order settings.

\textbf{Background Theories with Abnormality.}
Each instance specifies a fixed first-order theory $\Theta$ over an extended signature
$\Sigma_{\Ab}=\Sigma \cup \{\Ab\}$, where $\Ab(\cdot)$ is a unary abnormality predicate.
We use $\Theta$ to express default-like constraints that may be blocked by abnormality.
A common pattern is an implication schema of the form
\begin{equation}
\forall x\;\Big( \mathrm{Ante}(x)\;\wedge\;\neg \Ab(x)\;\rightarrow\;\mathrm{Cons}(x)\Big),
\label{eq:default-schema}
\end{equation}
where $\mathrm{Ante}$ and $\mathrm{Cons}$ are first-order formulas over $\Sigma$ (often involving
existentials over $R,S$).
Intuitively, elements satisfying the antecedent are expected to satisfy the consequent \emph{unless}
they are declared abnormal.

\textbf{Hypotheses as Definitions of Abnormality.}
A model outputs a single first-order formula $\alpha(x)$ (one free variable) over a restricted
set of allowed predicates (specified per instance).
We treat the prediction as a meta-level definition of abnormality:
\begin{equation}
\Ab := \alpha .
\label{eq:abd-def}
\end{equation}
Equivalently, $\Theta[\Ab \mapsto \alpha]$ denotes the repaired theory obtained by
substituting the instantiated formula $\alpha(t)$ for each atom $\Ab(t)$ in $\Theta$.
Example: under T1, suppose two prompt worlds contain violating objects $a$ and $c$, and in each world the violating object also has an $S$-successor. If $S$ is allowed in the hypothesis language, a shared rule such as $\alpha(x):=\exists z\,S(x,z)$ repairs both worlds jointly, illustrating that the task is to synthesize one shared rule, not one rule per world.

\textbf{Symbolic Output Language.}
To support exact evaluation, $\alpha(x)$ is produced in a simple S-expression syntax.
This syntax provides a canonical I/O format: prefix notation avoids precedence and associativity ambiguity and is robust inside single-line JSON outputs.
The grammar admits boolean connectives (\texttt{and}, \texttt{or}, \texttt{not}, \texttt{implies}),
first-order quantifiers (\texttt{forall}, \texttt{exists}), and atomic predicates from the allowed signature.
We use AST size, or formula size, to mean the number of nodes in the parsed formula tree of $\alpha$: each operator, predicate, variable, and constant counts as one node, and each quantifier contributes one node for the binder plus one for its bound variable.

\textbf{Unknown Atoms and Completions.}
In partial worlds, each ground atom of $R$ and $S$ is either \emph{known true}, \emph{known false}, or
\emph{unknown}.
Let $\Omega_W$ be the set of unknown atoms in $W$.
A \emph{completion} $\mathcal{C}$ assigns a truth value to every atom in $\Omega_W$, yielding a fully
specified world $W^{\mathcal{C}}$.
We write $\mathrm{KnownFacts}(W)$ for the conjunction of all observed positive facts and observed negative facts in $W$
(treating atoms not listed as true or unknown as observed false).
In fully observed worlds, $\mathrm{KnownFacts}(W)$ coincides with $\mathrm{Facts}(W)$.
For a completion $\mathcal{C}$, we take $\mathrm{Facts}(W^{\mathcal{C}})$ to be $\mathrm{KnownFacts}(W)$ together with the literals
that fix each atom in $\Omega_W$ according to $\mathcal{C}$.

\textbf{Finite Grounding and Solver Checks.}
All domains are finite, so satisfiability of $\Theta$ under a predicted $\alpha$ can be checked by grounding
quantifiers over $D_W$ and translating the result to SMT.
In partial worlds, unknown atoms in $\Omega_W$ are represented as free Boolean variables in the SMT encoding,
so we can quantify over completions by changing which constraints are asserted.
Existential-completion validity (ABD-Partial) reduces to a satisfiability check with unknowns left free,
whereas skeptical validity (ABD-Skeptical) reduces to the absence of a counterexample completion (checked as a single
UNSAT query that asks whether there exists a completion consistent with $\mathrm{KnownFacts}(W)$ under which some grounded axiom fails
in the repaired theory $\Theta[\Ab \mapsto \alpha]$).
Because costs are cardinalities over a finite domain, we can compute parsimony objectives exactly (or with
thresholded SAT checks) using Z3 \cite{demoura2008z3}.
Throughout, Z3 is the exact verifier and cost evaluator for these finite grounded instances.

We formalize the existential and universal completion semantics as synthesis problems in Section~\ref{sec:abduction}.

\section{Default--Exception Abduction Tasks}
\label{sec:abduction}

We now define the three ABD task variants as formal synthesis problems.
Fix an instance $\mathcal{I}$ with background theory $\Theta$ and prompt worlds
$\mathcal{W}_{\mathrm{p}}=\{W_1,\ldots,W_k\}$.
The learner outputs $\alpha(x)$ and is evaluated by (i) validity and (ii) parsimony.

\subsection{ABD-Full: Abduction Under Full Observation}
\label{sec:abd-full}

In ABD-Full, each world $W$ provides a complete interpretation of $P,Q,R,S$ under a closed-world assumption
for the given domain (any unlisted atom is false).
We write $\mathrm{SAT}(\varphi)$ to mean that the finite grounded formula $\varphi$ is satisfiable.
A hypothesis $\alpha$ is \emph{valid} on a world $W$ if the repaired theory is satisfiable:
\begin{multline}
W \models_{\mathrm{ABD}} \alpha \;\Longleftrightarrow \\
\mathrm{SAT}\big(\Theta[\Ab \mapsto \alpha] \wedge \mathrm{Facts}(W)\big).
\end{multline}
An $\alpha$ solves an instance iff it is valid on \emph{all} prompt worlds.

\textbf{Cost.}
For a valid hypothesis, we define the per-world abnormality count
\(
\mathrm{cost}_W(\alpha)=|\{a\in D_W : W\models \alpha[a]\}|.
\)
The instance-level cost is the sum across worlds:
\begin{equation}
\mathrm{Cost}(\alpha;\mathcal{W}_{\mathrm{p}})=\sum_{W\in\mathcal{W}_{\mathrm{p}}}\mathrm{cost}_W(\alpha).
\end{equation}

\textbf{Lower-Bound Baseline and Gap.}
To measure parsimony, we compare a model's cost to a solver-computed lower bound obtained by
allowing $\Ab$ to be assigned freely (not necessarily definable by a single $\alpha$).
Concretely, for each world we compute
\begin{equation}
\begin{aligned}
\mathrm{OptCost}(W)= {} & \min_{A\subseteq D_W}\;|A| \\
& \text{s.t.}\;\mathrm{SAT}\big(\Theta[\Ab \mapsto A] \wedge \mathrm{Facts}(W)\big),
\end{aligned}
\end{equation}
where $\Theta[\Ab \mapsto A]$ means that each grounded atom $\Ab(a)$ is set true iff $a \in A$,
and define $\mathrm{OptCost}(\mathcal{W}_{\mathrm{p}})=\sum_{W\in\mathcal{W}_{\mathrm{p}}}\mathrm{OptCost}(W)$.
We report the \emph{gap}
\(
\mathrm{Gap}(\alpha)=\mathrm{Cost}(\alpha;\mathcal{W}_{\mathrm{p}})-\mathrm{OptCost}(\mathcal{W}_{\mathrm{p}}).
\)
Gap is defined only for valid hypotheses; invalid outputs contribute to validity/error rates but are excluded from gap averages.
Because $\mathrm{OptCost}$ relaxes the single-formula constraint and allows $\Ab$
to vary freely by world, it lower-bounds the best \emph{definable} cost; thus
$\mathrm{Gap}(\alpha)$ conservatively overestimates suboptimality. On small domains 
the bound can often be matched by large extensional case-splits, 
motivating our AST- and holdout-based diagnostics.

This baseline is nevertheless solver-verifiable, inexpensive compared to full joint optimization,
and empirically discriminative across models.

\subsection{ABD-Partial: Existential Completion}
\label{sec:abd-partial}

ABD-Partial introduces missing information about the observed predicates (Section~\ref{sec:setup}).
Recall that $\Omega_W$ is the set of unknown atoms in $W$ and $W^{\mathcal{C}}$ denotes the completed world under completion $\mathcal{C}$.

\textbf{Validity Under Existential Completion.}
A hypothesis $\alpha$ is valid on $W$ if there exists a completion that makes the repaired theory satisfiable:
\begin{multline}
W \models_{\exists\mathrm{comp}} \alpha \;\Longleftrightarrow \\
\exists \mathcal{C}: \mathrm{SAT}\big(\Theta[\Ab \mapsto \alpha] \wedge \mathrm{Facts}(W^{\mathcal{C}})\big).
\end{multline}

\textbf{Cost Optimized Over Completions.}
Because $\alpha$ may mention unknown relations, its extension can depend on the completion.
We define cost as the minimum abnormality count over completions that satisfy the repaired theory:
\begin{multline}
\mathrm{cost}_W^{\exists}(\alpha)=
\min_{\mathcal{C}}\;|\{a\in D_W : W^{\mathcal{C}}\models \alpha[a]\}| \\
\text{s.t.}\;
\mathrm{SAT}\!\big(\Theta[\Ab \mapsto \alpha] \wedge \mathrm{Facts}(W^{\mathcal{C}})\big).
\label{eq:abd_partial_cost}
\end{multline}
Instance cost sums these per-world values, and gaps are defined relative to a completion-aware lower bound
that also leaves $\Ab$ unconstrained:
\begin{equation}
\begin{aligned}
\mathrm{OptCost}^{\exists}(W)= {} &
\min_{\mathcal{C},\,A\subseteq D_W}\;|A| \\
& \text{s.t.}\;\mathrm{SAT}\big(\Theta[\Ab \mapsto A] \wedge \mathrm{Facts}(W^{\mathcal{C}})\big).
\end{aligned}
\end{equation}
As in ABD-Full, the free-$\Ab$ lower bound is conservative (Section~\ref{sec:abd-full}).

\subsection{ABD-Skeptical: Universal Completion}
\label{sec:abd-skeptical}

ABD-Skeptical retains partial observation but treats unknown atoms conservatively.
Intuitively, a hypothesis should repair the default theory not only for \emph{some} completion of missing facts,
but for \emph{all} completions consistent with what is observed.

\textbf{Validity Under Universal Completion.}
A hypothesis $\alpha$ is \emph{skeptically valid} on $W$ if the repaired theory is satisfiable for every completion:
\begin{multline}
W \models_{\forall\mathrm{comp}} \alpha \;\Longleftrightarrow \\
\forall \mathcal{C}:\;
\mathrm{SAT}\big(\Theta[\Ab \mapsto \alpha] \wedge \mathrm{Facts}(W^{\mathcal{C}})\big).
\end{multline}
An $\alpha$ solves an instance iff it is valid on \emph{all} prompt worlds under this universal-completion semantics.

\textbf{Worst-Case Cost Over Completions.}
For a skeptically valid hypothesis, the per-world cost is the \emph{worst-case} abnormality count
over completions:
\begin{equation}
\mathrm{cost}_W^{\forall}(\alpha)=
\max_{\mathcal{C}}\;|\{a\in D_W : W^{\mathcal{C}}\models \alpha[a]\}|.
\end{equation}
The instance-level cost is the sum across prompt worlds:
\begin{equation}
\mathrm{Cost}^{\forall}(\alpha;\mathcal{W}_{\mathrm{p}})=\sum_{W\in\mathcal{W}_{\mathrm{p}}}\mathrm{cost}_W^{\forall}(\alpha).
\end{equation}

\textbf{Lower-Bound Baseline and Gap.}
We lift the free-$\Ab$ lower bound (Section~\ref{sec:abd-full}) to the skeptical setting by taking the worst case over completions:
\begin{equation}
\mathrm{OptCost}^{\forall}(W)=\max_{\mathcal{C}}\;\mathrm{OptCost}(W^{\mathcal{C}}),
\end{equation}
and define $\mathrm{OptCost}^{\forall}(\mathcal{W}_{\mathrm{p}})=\sum_{W\in\mathcal{W}_{\mathrm{p}}}\mathrm{OptCost}^{\forall}(W)$.
We report the skeptical gap
\begin{equation}
\mathrm{Gap}^{\forall}(\alpha)=\mathrm{Cost}^{\forall}(\alpha;\mathcal{W}_{\mathrm{p}})-\mathrm{OptCost}^{\forall}(\mathcal{W}_{\mathrm{p}}).
\end{equation}
This baseline is again conservative in the sense of Section~\ref{sec:abd-full}.

\subsection{Planted Generator References and Holdout Evaluation}
\label{sec:abd-metrics}

\textbf{Planted Generator References.}
Many instances are generated by selecting a planted generator reference abnormality rule $\alpha^\star$ from a controlled
template family and constructing worlds for which $\alpha^\star$ is valid and nontrivial.
We treat $\alpha^\star$ as a \emph{generator artifact}, not as the unique correct answer:
models may find different valid rules with lower or higher cost.

\textbf{Reference-Relative Cost and ``Beating the Reference.''}
To aid analysis and to detect shortcut hypotheses, we report the gap to the planted generator reference:
\(
\mathrm{Gap}_{\mathrm{ref}}(\alpha)=\mathrm{Cost}(\alpha)-\mathrm{Cost}(\alpha^\star).
\)
Negative values indicate that the model found a rule with fewer exceptions than the planted one.
We treat such cases as diagnostics that the planted generator reference is an anchor rather than an optimum.

\textbf{Size-Conditioned Metrics.}
A model can drive cost down by producing large disjunctive case splits tailored to the finite worlds.
To separate such behavior from compact reasoning, we report size-conditioned metrics that group formulas by AST size into bins (Table~\ref{tab:abd_complexity_by_scenario}).

\textbf{Holdout Worlds.}
To probe generalization, each instance also has $k{=}5$ \emph{holdout worlds} pre-generated from the same world sampler and theory parameters as the prompt worlds.
Unlike prompt worlds, holdout worlds satisfy the same per-world acceptance checks but are \emph{not} adversarially hardened against competitor formulas.
Holdout is therefore an \emph{out-of-filter distribution refresh}: matched in sampling distribution but not adversarially hardened.
Holdout worlds are pre-generated and cached for reproducibility.
We report holdout validity and cost gaps, as well as $\Delta\text{Gap}=\text{H-Gap}-\text{P-Gap}$ (the gap degradation from prompt worlds to holdout).
For ABD-Skeptical, holdouts additionally require that the planted generator reference satisfies universal-completion validity.
Prompt and holdout distributions remain closely matched in world counts and per-world reference costs.

\section{Dataset Generation}
\label{sec:generation}

This paper emphasizes \emph{controllable difficulty}: we want instances where simple exception rules fail,
multiple worlds jointly constrain the solution, and models cannot rely on a single brittle shortcut.
We therefore generate instances using three ingredients:

\textbf{(i) Controlled Hypothesis Families.}
The planted generator reference is sampled from a theory-compatible template library of S-expression formulas.
The library is organized by quantifier-depth tier: Tier~0 propositional unary formulas (QD~0), Tier~1 single-quantifier formulas (QD~1), and curated Tiers~2--3 nested-quantifier formulas, including family-specific variants for richer allowed signatures.
For each theory, we retain only templates whose predicates lie in the allowed hypothesis signature, so difficulty is controlled jointly by tier and theory-specific predicate scope.
A diversity tracker also limits repeated reuse of the same template within a theory$\times$difficulty bucket.

\textbf{(ii) Multi-World Filtering and Parsimony Floors.}
Candidate prompt-world sets are filtered both per world and jointly across worlds.
Per world, the planted generator reference must be valid, require at least one abnormal object ($\mathrm{OptCost}\ge 1$), remain near-optimal (reference cost $\le \mathrm{OptCost}+1$), and keep abnormalities sparse ($\mathrm{OptCost}/|D_W|\le 0.20$).
Aggregate difficulty refers to the corresponding multi-world totals, especially the summed lower-bound cost and summed reference cost across the prompt worlds, so that we reject instances whose joint repair problem is still too easy even if each world alone is nontrivial.

\textbf{(iii) CEGIS-Like Elimination of Shortcut Competitors.}
We then harden the prompt worlds against shortcut explanations.
For each instance we build a theory-compatible competitor pool from simple Tier~0/1 formulas, mined shortcut formulas from prior model outputs, and a small set of mutants of the planted generator reference.
A competitor is called a survivor if it remains valid on every current prompt world and stays within a small total-cost margin of the planted reference (typically~2).
When survivors remain, we add adversarial worlds that invalidate at least one survivor or make it sufficiently more costly, and accept the instance only when no survivors remain; otherwise the instance is rejected at the world-budget limit.

Table~\ref{tab:dataset_summary} summarizes the released benchmark.

\begin{table}[t]
\centering
\caption{\textbf{Abduction Benchmark Summary.} (a) Dataset statistics by scenario. (b) Default theories used in the benchmark; each theory defines when $\Ab(x)$ blocks a default conclusion.}
\label{tab:dataset_summary}
\footnotesize
\setlength{\tabcolsep}{3pt}

\begin{tabular}{@{}lrrrcl@{}}
\toprule
\multicolumn{6}{@{}l}{\textbf{(a) Dataset Statistics}} \\
\midrule
Scenario & N & Prompt & Holdout & $|D|$ & Theories \\
\midrule
FULL & 195 & 9.0 & 5.0 & 9--11 & T1, T2, T3, T4, T5 \\
PARTIAL & 243 & 8.3 & 5.0 & 9--11 & T1, T2, T3, T4, T5 \\
SKEPTICAL & 162 & 6.0 & 5.0 & 10--12 & \makecell[l]{T1, T2, T3, T4, T5,\\T6, T7} \\
\midrule
\textbf{Total} & \textbf{600} & & & & \\
\bottomrule
\end{tabular}

\vspace{1em}

\footnotesize
\begin{tabular}{@{}cp{5.8cm}@{}}
\toprule
\multicolumn{2}{@{}l}{\textbf{(b) Default Theories}: $\phi(x) \land \neg\Ab(x) \to \psi(x)$} \\
\midrule
ID & Antecedent $\to$ Consequent \\
\midrule
T1 & $\exists y(R(x,y) \land P(y))$ $\to$ $Q(x)$ \\
T2 & $\exists y(R(x,y) \land P(y))$ $\to$ $\exists z(S(x,z) \land Q(z))$ \\
T3 & $\exists y(S(x,y) \land P(y))$ $\to$ $\exists z(R(x,z) \land Q(z))$ \\
T4 & $\exists y(R(x,y) \land P(y))$ $\to$ $\exists z(S(x,z) \land \forall w(R(z,w) \to P(w)))$ \\
T5 & $\exists y(R(x,y) \land P(y))$ $\to$ $\forall z(S(x,z) \to Q(z))$ \\
T6 & $P(x)$ $\to$ $\exists y(R(x,y))$ \\
T7 & $P(x)$ $\to$ $\forall y(R(x,y) \to Q(y))$ \\
\bottomrule
\end{tabular}
\end{table}

\section{Evaluation and Experiments}
\label{sec:experiments}

\begin{table*}[t]
\centering
\caption{\textbf{Abduction Prompt-Set Performance.} Models are grouped by overall repaired prompt validity into high ($>90\%$), intermediate, and low ($<50\%$) bands; within each band, rows are ordered by overall Gap. Percentages are benchmark instances, not worlds. PV\% = prompt-valid after conservative suffix repair; PSV\% = strict prompt-validity without repair; AST = mean formula size among prompt-valid formulas; Gap = normalized extra exceptions above the solver lower bound; GRef = normalized extra exceptions vs. the planted generator reference. Gap and GRef average over prompt-valid formulas only. Lower is better for AST, Gap, and GRef. Best values bold.}
\label{tab:abd_main_train}
\footnotesize
\setlength{\tabcolsep}{1.2pt}
\begin{tabular}{@{}l rrrrr rrrrr rrrrr rrrrr@{}}
\toprule
 & \multicolumn{5}{c}{FULL} & \multicolumn{5}{c}{PARTIAL} & \multicolumn{5}{c}{SKEPTICAL} & \multicolumn{5}{c}{Overall} \\
\cmidrule(lr){2-6} \cmidrule(lr){7-11} \cmidrule(lr){12-16} \cmidrule(lr){17-21}
Model & PV\% & PSV\% & AST & Gap & GRef & PV\% & PSV\% & AST & Gap & GRef & PV\% & PSV\% & AST & Gap & GRef & PV\% & PSV\% & AST & Gap & GRef \\
\midrule
\multicolumn{21}{@{}l}{\textit{High Validity ($>90\%$)}} \\
Opus-4.6 & \textbf{100} & \textbf{100} & 12.4 & 1.26 & 0.74 & \textbf{98} & \textbf{98} & 15.9 & 1.02 & 0.55 & 99 & 99 & 15.8 & 0.86 & 0.17 & \textbf{99} & \textbf{99} & 14.7 & 1.05 & 0.51 \\
GPT-5.2 & 85 & 84 & 36.6 & 0.93 & 0.41 & 92 & 92 & 16.3 & 1.30 & 0.84 & \textbf{100} & \textbf{100} & 15.7 & 1.49 & 0.80 & 92 & 92 & 22.0 & 1.25 & 0.70 \\
Gemini-3.1 & 99 & 99 & 12.5 & 1.27 & 0.76 & 97 & 97 & 11.4 & 1.40 & 0.93 & 98 & 98 & 13.5 & 1.18 & 0.50 & 98 & 98 & 12.3 & 1.30 & 0.76 \\
Grok4.1f & 96 & 94 & 12.8 & 1.32 & 0.80 & 95 & 95 & 9.0 & 1.69 & 1.23 & 96 & 96 & 11.8 & 1.46 & 0.82 & 95 & 95 & 11.0 & 1.51 & 0.98 \\
DSR & 96 & 94 & 11.9 & 1.42 & 0.90 & 96 & 96 & 9.4 & 1.54 & 1.08 & 96 & 93 & 11.7 & 1.69 & 1.01 & 96 & 95 & 10.8 & 1.55 & 1.00 \\
\midrule
\multicolumn{21}{@{}l}{\textit{Intermediate Validity}} \\
GPT-5.4 & 76 & 68 & 106.1 & \textbf{0.13} & \textbf{-0.38} & 86 & 81 & 61.5 & \textbf{0.65} & \textbf{0.18} & 94 & 92 & 32.6 & \textbf{0.37} & \textbf{-0.28} & 85 & 80 & 65.9 & \textbf{0.42} & \textbf{-0.12} \\
Grok4 & 87 & 87 & 26.2 & 0.90 & 0.39 & 92 & 92 & 10.5 & 1.55 & 1.09 & 89 & 89 & 16.8 & 1.08 & 0.40 & 89 & 89 & 17.1 & 1.21 & 0.67 \\
Gemini-3 & 77 & 77 & 18.1 & 1.23 & 0.72 & 65 & 65 & 16.5 & 1.22 & 0.79 & 93 & 93 & 15.3 & 1.25 & 0.59 & 77 & 77 & 16.6 & 1.23 & 0.70 \\
Kimi-K2t & 60 & 57 & 13.3 & 1.45 & 0.93 & 73 & 72 & 10.3 & 1.68 & 1.21 & 83 & 80 & 10.3 & 2.41 & 1.74 & 72 & 70 & 11.1 & 1.84 & 1.30 \\
\midrule
\multicolumn{21}{@{}l}{\textit{Low Validity ($<50\%$)}} \\
Hermes4 & 3 & 3 & \textbf{8.2} & 2.85 & 2.31 & 1 & 1 & \textbf{8.0} & 1.78 & 1.44 & 4 & 4 & 6.5 & 3.11 & 2.56 & 2 & 2 & 7.4 & 2.80 & 2.29 \\
GPT-4o & 9 & 9 & 8.4 & 3.05 & 2.59 & 23 & 23 & 8.4 & 3.18 & 2.68 & 29 & 29 & \textbf{4.7} & 2.87 & 2.32 & 20 & 20 & \textbf{6.9} & 3.04 & 2.52 \\
\bottomrule
\end{tabular}
\end{table*}
\begin{table*}[t]
\centering
\caption{\textbf{Abduction Prompt-Set Performance by Theory.} PV\% = prompt-validity after conservative suffix repair; GRef = normalized extra exceptions vs. the planted generator reference, averaged over prompt-valid formulas only. T1--T5 aggregate across all scenarios; T6--T7 are ABD-Skeptical only. Lower is better for GRef. Best values bold.}
\label{tab:abd_theory_breakdown}
\footnotesize
\setlength{\tabcolsep}{1.0pt}
\begin{tabular}{@{}lrrrrrrrrrrrrrrrrrrrrrr@{}}
\toprule
 & \multicolumn{2}{c}{Opus-4.6} & \multicolumn{2}{c}{GPT-5.2} & \multicolumn{2}{c}{Gemini-3.1} & \multicolumn{2}{c}{Grok4.1f} & \multicolumn{2}{c}{DSR} & \multicolumn{2}{c}{GPT-5.4} & \multicolumn{2}{c}{Grok4} & \multicolumn{2}{c}{Gemini-3} & \multicolumn{2}{c}{Kimi-K2t} & \multicolumn{2}{c}{Hermes4} & \multicolumn{2}{c}{GPT-4o} \\
\cmidrule(lr){2-3} \cmidrule(lr){4-5} \cmidrule(lr){6-7} \cmidrule(lr){8-9} \cmidrule(lr){10-11} \cmidrule(lr){12-13} \cmidrule(lr){14-15} \cmidrule(lr){16-17} \cmidrule(lr){18-19} \cmidrule(lr){20-21} \cmidrule(lr){22-23}
Theory & PV\% & GRef & PV\% & GRef & PV\% & GRef & PV\% & GRef & PV\% & GRef & PV\% & GRef & PV\% & GRef & PV\% & GRef & PV\% & GRef & PV\% & GRef & PV\% & GRef \\
\midrule
T1 & \textbf{100} & 0.81 & 97 & 1.10 & 97 & 1.03 & 97 & 1.46 & 95 & 1.46 & 95 & \textbf{0.07} & 92 & 0.95 & 97 & 0.99 & 76 & 1.97 & 3 & 4.00 & 16 & 5.29 \\
T2 & \textbf{99} & 0.74 & 94 & 0.84 & \textbf{99} & 0.93 & 98 & 1.05 & 96 & 1.10 & 82 & \textbf{0.05} & 91 & 0.76 & 81 & 0.71 & 75 & 1.33 & 2 & 1.01 & 16 & 2.00 \\
T3 & \textbf{99} & 0.71 & 88 & 0.67 & 98 & 0.71 & 97 & 1.02 & 97 & 0.97 & 82 & \textbf{-0.14} & 90 & 0.74 & 70 & 0.63 & 74 & 1.36 & 1 & 2.00 & 28 & 1.25 \\
T4 & \textbf{99} & 0.46 & 88 & 0.78 & 97 & 0.89 & 91 & 1.12 & 95 & 1.13 & 86 & \textbf{-0.18} & 86 & 0.71 & 60 & 0.86 & 59 & 1.44 & 0 & --- & 14 & 4.66 \\
T5 & 98 & 0.16 & 96 & 0.61 & 98 & 0.56 & 92 & 0.80 & \textbf{99} & 0.86 & 85 & \textbf{-0.21} & 90 & 0.62 & 92 & 0.53 & 82 & 0.99 & 8 & 2.82 & 5 & 5.17 \\
T6 & \textbf{100} & -0.88 & \textbf{100} & -0.88 & \textbf{100} & -0.81 & \textbf{100} & -0.61 & 93 & -0.65 & 93 & \textbf{-0.99} & 86 & -0.71 & 93 & -0.79 & 64 & -0.62 & 0 & --- & 93 & 1.88 \\
T7 & 96 & 0.03 & \textbf{100} & 0.24 & \textbf{100} & 0.24 & 96 & 0.64 & 92 & 0.60 & 96 & \textbf{-0.05} & 92 & 0.19 & 96 & 1.07 & 88 & 1.16 & 4 & 1.00 & 60 & 1.67 \\
\bottomrule
\end{tabular}
\end{table*}
\begin{table}[t]
\centering
\caption{\textbf{Generalization Drop Profile.} Aggregated across all scenarios. PV\%/HV\% are instance-level prompt/holdout validity; PGap/HGap are normalized per-world gaps on their respective valid subsets; $\Delta$Gap is computed on survivors valid on both prompt and holdout; AST = mean formula size among prompt-valid formulas. Best values bold.}
\label{tab:abd_holdout_summary}
\footnotesize
\setlength{\tabcolsep}{3pt}
\begin{tabular}{@{}lrrrrrr@{}}
\toprule
Model & PV\% & HV\% & PGap & HGap & $\Delta$Gap & AST \\
\midrule
Opus-4.6 & \textbf{98.8\%} & 62.2\% & 1.05 & 2.24 & +0.97 & 14.7 \\
GPT-5.2 & 92.2\% & 61.5\% & 1.25 & 2.45 & +0.85 & 22.0 \\
Gemini-3.1 & 98.0\% & 69.7\% & 1.30 & 2.37 & +0.98 & 12.3 \\
Grok4.1f & 95.2\% & \textbf{82.2\%} & 1.51 & 2.54 & +0.94 & 11.0 \\
DSR & 95.9\% & 80.6\% & 1.55 & 2.64 & +0.96 & 10.8 \\
GPT-5.4 & 85.2\% & 24.8\% & \textbf{0.42} & \textbf{1.37} & \textbf{+0.56} & 65.9 \\
Grok4 & 89.4\% & 59.1\% & 1.21 & 2.32 & +0.95 & 17.1 \\
Gemini-3 & 76.9\% & 49.6\% & 1.23 & 2.13 & +0.80 & 16.6 \\
Kimi-K2t & 71.5\% & 64.8\% & 1.84 & 2.85 & +0.93 & 11.1 \\
Hermes4 & 2.3\% & 3.0\% & 2.80 & 4.71 & +1.17 & 7.4 \\
GPT-4o & 19.8\% & 19.0\% & 3.04 & 4.01 & +0.99 & 6.9 \\
\bottomrule
\end{tabular}
\end{table}

\subsection{Prompting and Inference Setup}

We employ a \textbf{zero-shot} prompting strategy: no benchmark instances or planted generator references appear as in-context examples.
Each prompt consists of four components:
(1)~a \emph{system instruction} establishing the role (``expert in first-order logic and abductive reasoning'') and requiring JSON output;
(2)~a \emph{scenario-specific task description} that defines the observation semantics (closed-world, existential, or universal completion), the S-expression grammar for~$\alpha(x)$, the validity/parsimony/simplicity objectives, and illustrative grammar examples;
(3)~the \emph{instance specification}, comprising the allowed and forbidden predicates, the theory axioms, and the prompt worlds with their predicate interpretations (and, for ABD-Partial/Skeptical, unknown atoms);
(4)~an \emph{output format block} requesting a single JSON line with the S-expression formula and a natural-language description.
The prompt includes structured reasoning guidelines (``identify which objects must be abnormal; look for a pattern; verify against all worlds'') but instructs models to reason internally and emit only the final JSON answer.

All models are queried at temperature $T=0.1$ for near-deterministic outputs.
For models offering a thinking mode, we used the provider-default reasoning configuration with 64K token budget; other models were queried in standard generation mode. We do not tune few-shot exemplars, model-specific chain-of-thought scaffolds, or bespoke reasoning prompts.
Each of the 600 benchmark instances is presented independently (single-turn, no multi-round refinement).
All results correspond to API snapshots from January--March 2026 (through March 7 for GPT-5.4); model behavior may differ in future versions.

Each model outputs a candidate $\hat\alpha(x)$ in S-expression syntax.
We compute prompt validity and exception cost as defined in Section~\ref{sec:abduction}, with cost semantics varying by scenario.
All gap values reported are \emph{normalized} by the number of worlds to keep scores comparable across instances.
Accordingly, prompt-set Gap averages only over prompt-valid predictions, holdout Gap averages only over holdout-valid predictions, and $\Delta$Gap is computed only on survivors valid on both prompt and holdout worlds.
All validity percentages in the main tables are instance-level, not per-world.
Each evaluation record corresponds to one model call on one benchmark instance, so a prediction counts as prompt-valid only if the same shared formula $\hat\alpha$ is valid on every prompt world in that instance; holdout-valid is defined analogously over that instance's holdout worlds.
For overall model comparisons, we also estimate 95\% percentile bootstrap confidence intervals over benchmark instances (2{,}000 resamples, stratified by scenario), using the same instance-level metrics and survivor-conditioned definitions as in the main tables.
When the only parse failure is a suffix of missing right parentheses, we conservatively auto-close the formula at end of string, evaluate the repaired form, and report such cases separately from unrepaired parse errors.
Syntax failures are separated from semantic invalidity in our reporting: across the deduplicated evaluation cache, 1.3\% of outputs required conservative suffix auto-closing and 3.4\% remained unrepaired parse failures, whereas 21.7\% were well-formed but semantically invalid on the prompt worlds; among the strongest models, unrepaired syntax failures are near-zero, with GPT-5.4 standing out mainly in repair rate (6.7\% auto-closed outputs).

\subsection{Main Results}

Table~\ref{tab:abd_main_train} summarizes prompt-set performance across models along three axes:
(i) \emph{validity} (does the repaired theory have any model?),
(ii) \emph{syntactic compactness} (how large are the prompt-valid formulas?), and
(iii) \emph{cost parsimony} (how close is the exception count to a solver lower bound?).
The table reports repaired prompt validity (PV\%), strict non-repaired prompt validity (PSV\%), AST, Gap (vs.\ the relaxed free-$\Ab$ lower bound; see Section~\ref{sec:abd-full}), and GRef (vs.\ the planted generator reference).
Two patterns stand out.

Frontier models already separate into distinct performance profiles.
Opus-4.6, Gemini-3.1, DSR, and Grok4.1f combine high prompt validity with relatively compact valid formulas, while GPT-5.4 pushes Gap and GRef much lower only by accepting substantially lower validity and much larger formulas.
This contrast matters because formula size is not merely cosmetic: the largest-AST systems generalize substantially worse to holdout worlds, suggesting that some models reduce prompt cost through brittle case-splitting rather than portable repair rules.
Bootstrap uncertainty does not materially change these conclusions. GPT-5.4 remains a clear low-gap/low-validity outlier under 95\% instance-level bootstrap intervals, and Opus-4.6 remains the lowest-gap model within the $>90\%$-validity cluster; by contrast, fine-grained ranking among DSR, Grok4.1f, Opus-4.6, and Gemini-3.1 on holdout validity or survivor-conditioned $\Delta$Gap should be treated as near-tied.

\textbf{Cost Gap and Formula Size Separate Different Models.}
Among valid predictions, semantic cost and syntactic compactness are only partially aligned.
Among the strongest high-validity models, normalized gaps cluster around roughly 1.0--1.6 additional abnormalities beyond the lower bound, while mean AST among valid formulas stays in the low teens (10.8--14.7 for Opus-4.6, Gemini-3.1, DSR, and Grok4.1f).
GPT-5.4 is the main exception: it attains the best overall prompt Gap (0.42) and GRef (-0.12), but this cost advantage comes with lower validity (85.2\%) and by far the largest valid formulas (AST 65.9 overall).
A low cost gap is therefore not by itself a sufficient measure of parsimony.
Interpreting the normalized scale: with $\approx 6$--$10$ prompt worlds per instance (Table~\ref{tab:dataset_summary}), a gap of $\approx 1.1$ corresponds to roughly $7$--$11$ extra abnormal elements per instance.

Several models ``beat the reference'' (negative reference-relative gap) on a nontrivial fraction of instances.
Planted generator references are \emph{generator anchors}, not ground truth; beating the reference often correlates with larger AST size.
GPT-5.4 makes this especially clear: it beats the reference on 56.8\% of valid ABD-Full/ABD-Partial instances, but those reference-beating formulas have mean AST 90.1, versus 52.1 for GPT-5.2, 22.4 for Opus-4.6, and 19.5 for Gemini-3.1.
This signal is diagnostic, not evidence of ``winning'' the task: cost parsimony relative to the solver lower bound remains the primary semantic metric, while AST tracks compactness.

\subsection{Scenario and Theory Breakdown}

Table~\ref{tab:abd_main_train} stratifies results by observation regime (FULL, PARTIAL, SKEPTICAL), while Table~\ref{tab:abd_theory_breakdown} breaks down performance by background theory.
Note that T1--T5 appear in all three scenarios, so their rows aggregate across regimes; T6--T7 are ABD-Skeptical only.
They were added only to enrich the skeptical regime: T6 stresses existential witness existence when $R$ facts may be unknown, while T7 stresses worst-case universal counterexamples created by unknown $R$ facts.

\textbf{Partial Observation Trades Feasibility for Stability.}
ABD-Partial often increases feasibility (higher validity) for some models by allowing existential completions of unknown atoms, but it does not uniformly improve parsimony.
In fact, models can remain valid while drifting farther from the lower bound, reflecting that ``having more ways to make the world consistent'' does not automatically translate into \emph{finding} parsimonious repairs.

\textbf{Skeptical Completion Changes the Failure Mode.}
ABD-Skeptical requires hypotheses to remain valid under \emph{all} completions of unknown atoms, not just some favorable one.
Prompt validity under skeptical semantics is not uniformly lower---some models even achieve higher prompt validity in Skeptical than in Full (Table~\ref{tab:abd_main_train})---but the cost metric differs: models must minimize the worst-case exception count, which penalizes formulas that ``get lucky'' under particular completions.
The main consequence of robust semantics emerges on holdout (Section~\ref{sec:experiments}): the primary failure mode shifts from parsimony inflation (in ABD-Full/ABD-Partial) to \emph{validity brittleness}, where rules that satisfy universal-completion on prompt worlds often break on holdouts.

\textbf{Model Families Respond Differently to Missingness.}
A notable qualitative contrast in Table~\ref{tab:abd_main_train} is that some models benefit from ABD-Partial (higher validity), while others lose validity under partial observation.
This suggests that the partial setting genuinely probes reasoning about \emph{what could be true} rather than merely making the task easier.
Gemini-3.1 improves sharply over Gemini-3 in every regime, especially on validity.
DSR and Grok4.1f combine high validity with low-teens valid-formula AST and later emerge as the strongest holdout-generalization pair.
Kimi-K2t shows the opposite tradeoff from GPT-5.4: once it finds a prompt-valid rule, it is unusually robust on holdout, but its gaps remain large.
GPT-5.4 drives cost down with much larger repair policies, which rarely survive holdout unchanged.

\textbf{Theory Choice Changes Which Shortcuts Exist.}
Across theories (Table~\ref{tab:abd_theory_breakdown}), T5 is consistently easier, while T3/T4 exhibit larger spread across models.
A plausible explanation is that theories with stronger relational structure (e.g., nested quantification or additional dependencies in consequents) constrain the exception rule more tightly, reducing the effectiveness of compact heuristics.
This theory-conditioned variation is valuable: it yields a natural set of controlled ``difficulty knobs'' without changing the output language or evaluation protocol.
GPT-5.4 also attains the best GRef in every theory row without leading validity on any theory family, reinforcing that low prompt cost and robust validity are distinct capabilities.
Additionally, T6 exhibits negative GRef values for most frontier models (Table~\ref{tab:abd_theory_breakdown}), so T6 GRef should be read as a generator-reference diagnostic rather than absolute optimality.

\subsection{Holdout Generalization}

Prompt worlds are the adversarially hardened worlds shown in the instance; holdout worlds are fresh matched draws from the same sampler and theory parameters, subject to the same per-world acceptance checks but \emph{without} competitor elimination (Section~\ref{sec:abd-metrics}).
Holdout is therefore an out-of-filter distribution refresh, not an i.i.d.\ split.
Because prompt worlds are harder by construction, holdout validity can exceed prompt validity for some models; the primary signal is the change in parsimony among rules that survive on both sets.

Table~\ref{tab:abd_holdout_summary} shows the overall generalization pattern: across models, holdout gaps are substantially larger than prompt gaps, with $\Delta$Gap typically around $+1$ extra exception per world among survivor instances.

\begin{table*}[!t]
\centering
\caption{\textbf{Holdout Generalization by Scenario.} PV\%/HV\% are prompt/holdout validity; PGap/HGap are normalized per-world gaps on their respective valid subsets; $\Delta$Gap is computed on survivors valid on both prompt and holdout. Best HV\% and $\Delta$Gap per scenario bold.}
\label{tab:abd_holdout_by_scenario}
\footnotesize
\setlength{\tabcolsep}{3pt}
\begin{tabular}{@{}lrrrrrrrrrrrrrrr@{}}
\toprule
Model & \multicolumn{5}{c}{FULL} & \multicolumn{5}{c}{PARTIAL} & \multicolumn{5}{c}{SKEPTICAL} \\
\cmidrule(lr){2-6} \cmidrule(lr){7-11} \cmidrule(lr){12-16}
 & PV\% & HV\% & PGap & HGap & $\Delta$Gap & PV\% & HV\% & PGap & HGap & $\Delta$Gap & PV\% & HV\% & PGap & HGap & $\Delta$Gap \\
\midrule
Opus-4.6 & 100\% & 87\% & 1.26 & 2.66 & +1.34 & 98\% & 51\% & 1.02 & 2.09 & +0.92 & 99\% & 49\% & 0.86 & 1.59 & +0.25 \\
GPT-5.2 & 85\% & 54\% & 0.93 & 2.74 & +1.49 & 92\% & 68\% & 1.30 & 2.52 & +1.02 & 100\% & 61\% & 1.49 & 2.06 & \textbf{-0.02} \\
Gemini-3.1 & 99\% & 87\% & 1.27 & 2.60 & +1.35 & 97\% & 68\% & 1.40 & 2.44 & +1.00 & 98\% & 52\% & 1.18 & 1.80 & +0.21 \\
Grok4.1f & 96\% & \textbf{88\%} & 1.32 & 2.59 & +1.27 & 95\% & \textbf{85\%} & 1.69 & 2.85 & +1.14 & 96\% & 69\% & 1.46 & 1.83 & -0.02 \\
DSR & 96\% & 85\% & 1.42 & 2.86 & +1.42 & 96\% & \textbf{85\%} & 1.54 & 2.72 & +1.11 & 96\% & 70\% & 1.69 & 2.23 & +0.09 \\
GPT-5.4 & 76\% & 18\% & 0.13 & 1.07 & \textbf{+0.68} & 86\% & 25\% & 0.65 & 2.13 & +0.96 & 94\% & 33\% & 0.37 & 0.73 & +0.04 \\
Grok4 & 87\% & 61\% & 0.90 & 1.86 & +0.98 & 92\% & 72\% & 1.55 & 2.80 & +1.12 & 89\% & 38\% & 1.08 & 1.85 & +0.43 \\
Gemini-3 & 77\% & 60\% & 1.23 & 2.42 & +1.21 & 65\% & 40\% & 1.22 & 2.21 & +0.98 & 93\% & 52\% & 1.25 & 1.66 & +0.11 \\
Kimi-K2t & 60\% & 58\% & 1.45 & 2.89 & +1.45 & 73\% & 66\% & 1.68 & 2.87 & +1.16 & 83\% & \textbf{71\%} & 2.41 & 2.76 & +0.10 \\
Hermes4 & 3\% & 5\% & 2.85 & 5.51 & +1.40 & 1\% & 1\% & 1.78 & 3.00 & +1.22 & 4\% & 4\% & 3.11 & 4.07 & +0.96 \\
GPT-4o & 9\% & 17\% & 3.05 & 5.58 & +2.03 & 23\% & 16\% & 3.18 & 3.31 & \textbf{+0.86} & 29\% & 25\% & 2.87 & 3.38 & +0.83 \\
\bottomrule
\end{tabular}
\end{table*}
\begin{table*}[!b]
\centering
\caption{\textbf{Holdout Generalization by Formula Complexity.} H$|$P columns report holdout validity conditional on prompt validity (percent); $\Delta$ columns report survivor-conditioned $\Delta$Gap. Hermes4 and GPT-4o are omitted due to too few valid responses.}
\label{tab:abd_complexity_by_scenario}
\footnotesize
\setlength{\tabcolsep}{1.8pt}
\begin{tabular}{@{}lrrrrrrrrrrrrrrrrrr@{}}
\toprule
 & \multicolumn{6}{c}{FULL} & \multicolumn{6}{c}{PARTIAL} & \multicolumn{6}{c}{SKEPTICAL} \\
\cmidrule(lr){2-7} \cmidrule(lr){8-13} \cmidrule(lr){14-19}
 & \multicolumn{2}{c}{{[}0,15{)}} & \multicolumn{2}{c}{{[}15,30{)}} & \multicolumn{2}{c}{{[}30,$\infty${)}} & \multicolumn{2}{c}{{[}0,15{)}} & \multicolumn{2}{c}{{[}15,30{)}} & \multicolumn{2}{c}{{[}30,$\infty${)}} & \multicolumn{2}{c}{{[}0,15{)}} & \multicolumn{2}{c}{{[}15,30{)}} & \multicolumn{2}{c}{{[}30,$\infty${)}} \\
\cmidrule(lr){2-3} \cmidrule(lr){4-5} \cmidrule(lr){6-7} \cmidrule(lr){8-9} \cmidrule(lr){10-11} \cmidrule(lr){12-13} \cmidrule(lr){14-15} \cmidrule(lr){16-17} \cmidrule(lr){18-19}
Model & H$|$P & $\Delta$ & H$|$P & $\Delta$ & H$|$P & $\Delta$ & H$|$P & $\Delta$ & H$|$P & $\Delta$ & H$|$P & $\Delta$ & H$|$P & $\Delta$ & H$|$P & $\Delta$ & H$|$P & $\Delta$ \\
\midrule
Opus-4.6 & 98 & +1.5 & 62 & +0.6 & 50 & \textbf{0.0} & 80 & +1.1 & 30 & +0.5 & 0 & --- & 73 & +0.3 & 26 & +0.1 & 10 & +2.2 \\
GPT-5.2 & 97 & +1.7 & 65 & +0.9 & 17 & +0.8 & 89 & +1.1 & 42 & +0.7 & 12 & +0.9 & 88 & \textbf{-0.1} & 24 & +0.3 & 12 & +1.7 \\
Gemini-3.1 & 93 & +1.6 & 71 & +0.8 & \textbf{100} & +0.1 & 78 & \textbf{+1.0} & 38 & +0.7 & --- & --- & 73 & +0.2 & 18 & 0.0 & \textbf{33} & +0.5 \\
Grok4.1f & 99 & +1.4 & 64 & +0.7 & 89 & +0.1 & 93 & +1.2 & 33 & \textbf{-0.1} & 0 & --- & 87 & 0.0 & \textbf{31} & +0.2 & --- & --- \\
DSR & 96 & +1.6 & 57 & +0.7 & 75 & +0.1 & 89 & +1.1 & \textbf{70} & +1.2 & --- & --- & 86 & +0.1 & 26 & 0.0 & 0 & --- \\
GPT-5.4 & \textbf{100} & +1.9 & \textbf{77} & \textbf{+0.4} & 15 & +0.6 & 84 & \textbf{+1.0} & 32 & +1.0 & 9 & +0.8 & 67 & \textbf{-0.1} & 26 & \textbf{-0.1} & 16 & +0.7 \\
Grok4 & 89 & +1.5 & 69 & +0.7 & 50 & +0.8 & 90 & +1.1 & 17 & +1.2 & 0 & --- & 67 & +0.5 & 26 & +0.2 & 7 & \textbf{-1.0} \\
Gemini-3 & 93 & \textbf{+1.3} & 61 & +1.1 & 50 & +0.7 & 82 & +1.1 & 30 & +0.4 & \textbf{27} & \textbf{+0.7} & 77 & +0.2 & 23 & \textbf{-0.1} & 9 & -0.9 \\
Kimi-K2t & \textbf{100} & +1.5 & 73 & +0.5 & 70 & +1.8 & \textbf{95} & +1.2 & 35 & +0.9 & 0 & --- & \textbf{97} & +0.1 & 6 & +0.5 & 0 & --- \\
\bottomrule
\end{tabular}
\end{table*}

\textbf{Two Distinct Failure Modes.}
Table~\ref{tab:abd_holdout_summary} aggregates across all three scenarios and mixes two effects: (i) whether a rule stays satisfiable on holdouts (\emph{validity brittleness}), and (ii) how parsimonious it remains when satisfiable (\emph{parsimony inflation}).
Conditional holdout validity further disentangles these by reporting holdout validity conditioned on prompt validity.
Because $\Delta$Gap is computed only on the survivor intersection, a small $\Delta$Gap does not by itself imply strong generalization.
GPT-5.4 is the clearest example: it has the best overall $\Delta$Gap (+0.56) but only 28.5\% conditional holdout validity.
These two failure modes manifest differently across scenarios, as we show next.

\textbf{Scenario-Level Holdout Analysis.}
Table~\ref{tab:abd_holdout_by_scenario} breaks down holdout generalization by scenario, revealing a striking contrast.
In ABD-Full and ABD-Partial, the main failure mode is \emph{parsimony inflation}: Grok4.1f and DSR achieve the strongest holdout validity, but even they pay roughly one extra exception per world on fresh samples.
In ABD-Skeptical, the pattern differs: $\Delta$Gap is smaller, but holdout validity drops are more severe---many rules that satisfy universal-completion on prompt worlds fail outright on holdouts.
Kimi-K2t reaches the highest skeptical holdout validity (71\%) at substantially worse cost, while GPT-5.4 again shows that low $\Delta$Gap without survival is not enough (33\% holdout validity).
This suggests that robust semantics regularizes cost inflation but makes validity harder to preserve.

\subsection{Complexity and Generalization}

Table~\ref{tab:abd_complexity_by_scenario} relates formula size (AST bins: $<$15, 15--30, $\geq$30) to holdout degradation across all three tasks.
Very small formulas tend to underfit (high $\Delta$Gap), while very large formulas reintroduce brittleness via case-splitting.
Moderate-sized formulas generally degrade least.
On paired problems where multiple models produced valid formulas, shorter-or-reference-length formulas have much higher conditional holdout validity (85\% overall vs.\ 28\% for longer formulas), while longer formulas achieve smaller survivor-conditioned $\Delta$Gap (+0.52 vs.\ +0.95 overall).
Lower cost can therefore be purchased by brittle case-splitting.

\subsection{Discussion}
Models differ qualitatively in their performance profiles. Opus-4.6, Gemini-3.1, DSR, and Grok4.1f form a high-validity cluster; within that group, DSR and Grok4.1f form the strongest compact holdout-generalization pair, with average valid-formula AST near 11, while Opus-4.6 and Gemini-3.1 lead in prompt validity.
GPT-5.4 is the clearest low-gap outlier: it achieves the best prompt Gap and GRef, but does so with the largest valid formulas (AST 65.9 overall) and weak holdout survival (24.8\% overall; 28.5\% conditional on prompt validity).
Kimi-K2t shows nearly the opposite tradeoff, remaining compact and conditionally robust but with substantially larger gaps.
Taken together, the results support three conclusions.

\textbf{(i) ABD Is Not Solved.}
Among the high-validity systems, cost parsimony remains challenging: prompt-set gap stays around one extra abnormality per world beyond the solver lower bound, and this gap roughly doubles on holdouts in ABD-Full/Partial.

\textbf{(ii) Holdout Evaluation Across All Three Regimes Is Essential.}
Prompt metrics alone are misleading because prompt sets are adversarially filtered.
Holdouts reveal different patterns by scenario: parsimony inflation dominates in ABD-Full/Partial, while 
validity brittleness dominates in ABD-Skeptical, suggesting that robust semantics regularizes cost but makes validity harder to preserve.
They also show that low survivor-conditioned $\Delta$Gap is not sufficient: GPT-5.4 has the smallest overall $\Delta$Gap, yet only 28.5\% conditional holdout validity.

\textbf{(iii) Cost- and Size-Aware Reporting Is Necessary.}
Binary validity saturates; cost gaps reveal one axis of semantic parsimony; AST reveals whether that cost is achieved with compact formulas or with bloated case-splitting.
The strongest counterexample is GPT-5.4, whose best-in-paper gap scores come with AST 65.9 and weak holdout survival.
More broadly, longer-than-reference formulas attain lower $\Delta$Gap than shorter ones (+0.52 vs.\ +0.95) while collapsing holdout validity (28\% vs.\ 85\%).
Improved abductive reasoning therefore requires improvement in all three metrics: validity, cost gap, and formula size.

\section{Conclusion}
We introduced \textsc{ABD}, a solver-checkable benchmark for multi-world default--exception abduction over finite first-order worlds, with exact SMT-based evaluation under full, partial, and skeptical observation regimes. This lets us assess not only whether a model can synthesize a valid repair rule across prompt worlds, but also how that rule trades off parsimony, syntactic complexity, and generalization.

Across tasks, frontier models often produce \emph{valid} repairs on prompt worlds, but \textsc{ABD} remains unsaturated 
because \emph{validity}, \emph{cost parsimony}, and \emph{syntactic complexity} do not saturate together. Among the high-validity models, the normalized gap to the solver lower bound 
remains roughly $1.0$--$1.6$ extra abnormalities per world, and the more compact systems stay in the low teens of valid-formula AST, while GPT-5.4 reaches much lower prompt gaps only by using much larger formulas.

Holdout evaluation shows that generalization remains the key challenge, and that its failure mode depends on the completion semantics. In ABD-Full and ABD-Partial, the main failure is \emph{parsimony inflation}: even when a rule remains valid on holdout worlds, its cost gap usually increases by about one additional exception per world, indicating overfitting to the filtered prompt worlds.

In contrast, ABD-Skeptical induces a different generalization profile. For rules that remain skeptically valid, 
gap inflation is often smaller than in ABD-Full/ABD-Partial, consistent with the worst-case objective acting as an 
implicit regularizer on exception marking. However, the main failure becomes \emph{validity brittleness}: rules that 
satisfy the universal-completion criterion on prompt worlds frequently fail outright on holdouts. Taken together, these 
outcomes suggest that robust completion semantics shifts difficulty from ``how few exceptions can I mark?'' to ``can I 
preserve a universally valid repair policy under distribution refresh?''

Finally, complexity matters: longer-than-reference formulas have only 28\% holdout validity versus 85\% for shorter formulas, even though their $\Delta$Gap is lower.
\textsc{ABD} should therefore be read as a joint profile over validity, parsimony, and syntactic budget.

\textbf{Limitations.}
Domains are small to permit exact SMT verification, which also makes extensional case-splitting feasible; we therefore report size-conditioned and holdout metrics to diagnose memorization. Our free-$\Ab$ baseline relaxes the single-formula constraint, so gap-to-opt is conservative, and planted generator references are anchors rather than uniquely optimal targets.

\textbf{Software and Data.}
Code and data are available at \url{https://github.com/SerafimBatzoglou/concept-synth}, including benchmark files, prompts, cached outputs, evaluation scripts, and instructions to reproduce the reported tables and figures.

\section*{Acknowledgements}
This work used Anthropic Claude Code and OpenAI Codex for software development and editorial support. All scientific decisions, experiments, verification, and final writing were performed and validated by the author.

\clearpage
\onecolumn
\appendix

\renewcommand{\thetable}{\Alph{section}.\arabic{table}}
\renewcommand{\thefigure}{\Alph{section}.\arabic{figure}}
\makeatletter
\@addtoreset{table}{section}
\@addtoreset{figure}{section}
\makeatother
\section{Task Prompts}
\label{sec:prompts}

Each benchmark instance is presented to the model as a single user
message preceded by a fixed system prompt.  The system prompt is shared
across all three observation regimes (ABD-Full, ABD-Partial,
ABD-Skeptical).  The user message consists of:
\begin{enumerate}[nosep]
  \item A \textbf{task description} explaining the observation
        semantics, S-expression grammar, and
        validity/parsimony objectives;
  \item The \textbf{predicate scope}
        (allowed and forbidden predicates for the hypothesis);
  \item The \textbf{theory} (axioms in S-expression form);
  \item The \textbf{prompt worlds} (domain, predicate
        interpretations, and---for partial/skeptical---unknown atoms);
  \item An \textbf{output format} specification requesting JSON with
        the formula and a natural-language description.
\end{enumerate}
Below we reproduce the system prompt followed by one concrete example
prompt for each scenario, generated directly from the benchmark's
prompt-building code.

\subsection{System Prompt (All Scenarios)}
\label{sec:system-prompt}

The following system prompt is prepended to every API call,
regardless of scenario.

\begin{lstlisting}[style=prompt]
You are an expert in first-order logic and abductive reasoning. 
Your task is to find concise formulas that explain abnormal behavior in logical systems.
Always output valid JSON with the required fields.
\end{lstlisting}

\subsection{ABD-Full Example Prompt}
\label{sec:prompt-full}

The following prompt was generated from a benchmark instance with
theory \texttt{TH7} (T5) and 6 prompt worlds under the
\textbf{closed-world} observation regime.  All predicate truth values
are fully observed; unlisted atoms are false.

\begin{lstlisting}[style=prompt]
# First-Order Logic Abduction (Full Observation)

## Task Overview

You are given a **default reasoning theory** and several finite "worlds" with **full observation**:
- A set of theory axioms involving an abnormality predicate Ab(x)
- Multiple worlds, each with a finite domain and complete predicate interpretations
- Your task: find a formula alpha(x) that defines which objects are "abnormal"

Each world contains:
- A finite domain of objects (named a0, a1, a2, ...)
- Complete interpretations of predicates P(x), Q(x), R(x,y), S(x,y)

**Closed World Assumption**: Only the facts explicitly listed as TRUE are true. Any predicate application not listed should be assumed FALSE.

## Abnormality Semantics

The abnormality predicate Ab(x) is defined by your formula:
- **Ab(x) <-> alpha(x)**: An object a is abnormal if and only if alpha(a) evaluates to TRUE

The theory axioms use Ab(x) to express default rules with exceptions. For example:
- "P(x) AND NOTAb(x) -> Q(x)" means: normally, P-objects have Q, unless they are abnormal

**Your goal**: Find a formula alpha(x) such that:
1. **Validity**: When Ab(x) is defined as alpha(x), all theory axioms are satisfied in every prompt world
2. **Parsimony**: Among valid formulas, prefer those that minimize the number of abnormal objects

## Scoring / Tie-breaks (important)

- alpha(x) is a SINGLE global formula shared across all worlds.
- Primary objective: **Validity** (must satisfy all axioms in every prompt world).
- Secondary objective: **Parsimony** -- minimize the total number of abnormal objects across prompt worlds:
  cost = SUM_world |{ a in Domain_world : alpha(a) is TRUE }|
- Tertiary objective: **Simplicity** -- if multiple formulas have the same validity + cost, prefer smaller formulas (lower AST / fewer operators / fewer quantifiers).

## Output Format

You must output your formula in S-expression syntax. The grammar is:

```
alpha ::= (P x)              -- unary predicate applied to variable
    | (R x y)            -- binary predicate applied to two variables  
    | (= x y)            -- equality of two variables
    | (not alpha)            -- negation
    | (and alpha_1 alpha_2 ...)    -- conjunction (2 or more arguments)
    | (or alpha_1 alpha_2 ...)     -- disjunction (2 or more arguments)
    | (forall v alpha)       -- universal quantification
    | (exists v alpha)       -- existential quantification
```

**Important constraints:**
- Your formula must have exactly one free variable: x
- All other variables must be bound by quantifiers (forall or exists)
- Variable names should be: x (free), y, z, w (bound by quantifiers)
- **Do NOT mention Ab in your formula.**
- **Use ONLY the predicate symbols listed in AllowedAlphaPredicates for this instance.**
- **If your formula uses any predicate in ForbiddenAlphaPredicates, it is invalid.**
- **Do NOT use object names like a0, a1, ... inside alpha(x). Use variables only.**
- Prefer simpler formulas; aim for minimal abnormal set
- **No implication or biconditional**: Express these using other connectives:
  - "A implies B" should be written as `(or (not A) B)`
  - "A if and only if B" should be written as `(and (or (not A) B) (or (not B) A))`

## Examples of Valid Formulas

**Note**: Examples below are syntactic illustrations; only use predicates that appear in AllowedAlphaPredicates for THIS instance.

- `(P x)` -- "x is abnormal if x satisfies P"
- `(and (P x) (not (Q x)))` -- "x is abnormal if x has P but not Q"
- `(exists y (and (R x y) (not (P y))))` -- "x is abnormal if x has an R-successor without P"
- `(not (exists y (R x y)))` -- "x is abnormal if x has no R-successors"
- `(forall y (or (not (R x y)) (Q y)))` -- "x is abnormal if all R-successors of x have Q"

## Evaluation

Your formula will be evaluated on:
1. **Validity**: Does defining Ab(x) <-> alpha(x) satisfy all theory axioms in every world?
2. **Parsimony**: How many objects are marked as abnormal? (fewer is better)
3. **Generalization**: Does the formula also work on holdout worlds?

**Note**: A trivially true formula like `(or (P x) (not (P x)))` makes everything abnormal (valid but not parsimonious). A trivially false formula makes nothing abnormal (may violate axioms).

---

## Problem Instance


**AllowedAlphaPredicates**: ["P", "R"]
**ForbiddenAlphaPredicates**: ["Ab", "Q", "S"]

**Theory ID**: TH7

**Axioms**:
1. `(forall x (implies (and (exists y (and (R x y) (P y))) (not (Ab x))) (exists z (and (S x z) (Q z)))))`

## Prompt Worlds

### World W0
Domain: {a0, a1, a2, a3, a4, a5, a6, a7, a8, a9, a10}

**Predicates** (Closed World Assumption: unlisted atoms are false):
- P: {a7, a3, a0, a9}
- Q: {a0, a6, a8}
- R: {(a10, a9), (a8, a2), (a5, a8), (a0, a2), (a7, a9), (a3, a1), (a4, a4), (a0, a8), (a4, a2), (a2, a1), (a0, a10), (a3, a4), (a9, a9), (a1, a1), (a6, a4), (a4, a0), (a4, a1), (a6, a1), (a10, a0), (a9, a7), (a8, a10), (a6, a10), (a5, a1), (a1, a6), (a1, a8), (a6, a5), (a8, a1)}
- S: {(a4, a7), (a10, a7), (a3, a1), (a4, a3), (a0, a0), (a5, a0), (a4, a6), (a8, a3), (a10, a8), (a4, a5), (a2, a3), (a6, a2), (a9, a10), (a8, a9), (a2, a9), (a5, a5), (a7, a9), (a3, a7), (a5, a6)}

### World W1
Domain: {a0, a1, a2, a3, a4, a5, a6, a7, a8, a9, a10}

**Predicates** (Closed World Assumption: unlisted atoms are false):
- P: {a6, a1}
- Q: {a5, a10, a0, a2}
- R: {(a10, a7), (a4, a1), (a4, a9), (a9, a5), (a10, a8), (a6, a3), (a7, a7), (a10, a9), (a8, a6), (a4, a0), (a6, a9), (a2, a8), (a1, a1), (a2, a0), (a3, a10), (a7, a3), (a0, a5), (a3, a7), (a10, a5), (a0, a0)}
- S: {(a2, a3), (a10, a7), (a7, a1), (a4, a0), (a4, a7), (a5, a9), (a9, a5), (a2, a5), (a7, a9), (a7, a3), (a1, a1), (a6, a4), (a6, a2), (a6, a1), (a2, a1), (a1, a6), (a8, a4), (a7, a0), (a7, a4), (a8, a10)}

### World W2
Domain: {a0, a1, a2, a3, a4, a5, a6, a7, a8, a9}

**Predicates** (Closed World Assumption: unlisted atoms are false):
- P: {a5, a9}
- Q: {a4, a6, a7, a3}
- R: {(a6, a5), (a2, a7), (a4, a8), (a1, a4), (a0, a5), (a5, a5), (a7, a2), (a4, a3), (a3, a4), (a7, a3), (a6, a3), (a5, a7), (a8, a6), (a7, a5), (a9, a1), (a1, a0), (a6, a7), (a8, a1), (a3, a2), (a4, a1), (a0, a0)}
- S: {(a1, a3), (a2, a1), (a7, a9), (a6, a2), (a0, a5), (a2, a4), (a4, a7), (a5, a2), (a6, a7), (a2, a8), (a1, a8), (a0, a8), (a7, a4), (a8, a1), (a7, a7), (a5, a0), (a3, a8)}

### World W3
Domain: {a0, a1, a2, a3, a4, a5, a6, a7, a8, a9}

**Predicates** (Closed World Assumption: unlisted atoms are false):
- P: {a7, a1, a0}
- Q: {a5, a8}
- R: {(a1, a6), (a2, a6), (a8, a3), (a9, a4), (a8, a8), (a9, a6), (a6, a4), (a0, a0), (a3, a2), (a6, a2), (a3, a0), (a5, a1), (a3, a9), (a7, a5), (a7, a8), (a8, a9), (a0, a5), (a2, a3)}
- S: {(a6, a0), (a6, a3), (a4, a8), (a8, a1), (a6, a4), (a9, a8), (a9, a4), (a3, a7), (a0, a3), (a6, a1), (a0, a1), (a1, a5), (a5, a8), (a9, a3), (a3, a8)}

### World W4
Domain: {a0, a1, a2, a3, a4, a5, a6, a7, a8, a9}

**Predicates** (Closed World Assumption: unlisted atoms are false):
- P: {a1, a4, a5, a7}
- Q: {a7, a2, a9}
- R: {(a5, a8), (a1, a6), (a4, a5), (a8, a3), (a9, a9), (a8, a8), (a0, a0), (a5, a6), (a3, a4), (a2, a9), (a6, a9), (a7, a6), (a1, a0)}
- S: {(a7, a6), (a0, a0), (a9, a2), (a1, a8), (a1, a3), (a2, a1), (a2, a3), (a5, a8), (a3, a1), (a1, a4), (a4, a7)}

### World W5
Domain: {a0, a1, a2, a3, a4, a5, a6, a7, a8, a9}

**Predicates** (Closed World Assumption: unlisted atoms are false):
- P: {a8, a3, a7}
- Q: {a3, a6, a4}
- R: {(a8, a8), (a1, a0), (a9, a4), (a5, a8), (a9, a3), (a3, a5), (a0, a4), (a3, a6), (a8, a3), (a5, a1), (a3, a7), (a2, a4), (a3, a2), (a4, a0), (a5, a3), (a1, a3), (a7, a8), (a6, a4), (a9, a0), (a6, a5), (a1, a5)}
- S: {(a8, a8), (a3, a3), (a4, a9), (a7, a8), (a6, a3), (a6, a5), (a6, a6), (a8, a4), (a9, a4), (a9, a2), (a8, a3), (a1, a4), (a0, a5)}

---

## Your Task

Analyze the theory axioms and prompt worlds carefully. Find a formula alpha(x) that defines abnormality such that all axioms are satisfied.

**Solve method (reason internally; do NOT output your reasoning):**
1. Understand the theory axioms: What do they require? When might they be violated?
2. For each world, identify which objects would violate an axiom if they were NOT abnormal
3. Look for a pattern: What property characterizes objects that MUST be abnormal?
4. Check for parsimony: Is there a simpler formula that still satisfies all axioms?
5. Formulate your hypothesis as an S-expression formula
6. **Verify**: For each world, check that defining Ab(x) <-> alpha(x) makes all axioms true

**Reminder**: Validity comes first. Only after you have a valid alpha(x), optimize parsimony (fewer abnormal objects), then simplify the formula.

**Example reasoning** (assumes predicates shown are allowed for alpha in this instance):
- Theory: "P(x) AND NOTAb(x) -> Q(x)" (normally, P-objects have Q)
- World: P(a0)=T, Q(a0)=F, P(a1)=T, Q(a1)=T
- Analysis: a0 has P but not Q, so if a0 is not abnormal, the axiom fails
- Result: alpha(x) = (and (P x) (not (Q x))) makes a0 abnormal, satisfying the axiom

**Important**: 
- Your formula must make the axioms true in ALL prompt worlds
- Minimize the number of abnormal objects when possible
- Do NOT mention Ab in your formula
- Use ONLY predicates from AllowedAlphaPredicates; using ForbiddenAlphaPredicates is invalid
- Do NOT use object names like a0, a1, ... inside alpha(x); use variables only

---

## Output

Your output must be exactly ONE LINE containing ONLY valid JSON with exactly these keys:
- "formula": a single S-expression formula string with one free variable x
- "description": a short plain-English description (one sentence)

Do NOT include code fences, prefixes, or any other text.

Example of correct output:
{"formula":"(and (P x) (not (Q x)))","description":"Objects with P but not Q are abnormal."}
\end{lstlisting}

\subsection{ABD-Partial Example Prompt}
\label{sec:prompt-partial}

The following prompt was generated from a benchmark instance with
theory \texttt{TH11} (T6) and 6 prompt worlds under the
\textbf{existential-completion} (partial observation) regime.  Known
true facts and unknown atoms are listed explicitly; unlisted atoms
that are not unknown are known false.  The model must find a formula
valid under \emph{at least one} completion of unknowns per world.

\begin{lstlisting}[style=prompt]
# First-Order Logic Abduction (Partial Observation)

## Task Overview

You are given a **default reasoning theory** and several finite "worlds" with **partial observation**:
- A set of theory axioms involving an abnormality predicate Ab(x)
- Multiple worlds, each with a finite domain and partially observed predicate interpretations
- Some predicate facts are **unknown** (not observed)
- Your task: find a formula alpha(x) that defines which objects are "abnormal"

Each world contains:
- A finite domain of objects (named a0, a1, a2, ...)
- Known facts: predicates whose truth values have been observed
- Unknown atoms: predicates whose truth values are hidden

**Observation Rules**:
- Atoms listed under "Known Facts" with TRUE values are TRUE
- Atoms listed under "Unknown Atoms" have unknown truth values
- Any atom NOT listed as TRUE or Unknown is FALSE

## Abnormality Semantics

The abnormality predicate Ab(x) is defined by your formula:
- **Ab(x) <-> alpha(x)**: An object a is abnormal if and only if alpha(a) evaluates to TRUE

The theory axioms use Ab(x) to express default rules with exceptions.

**Your goal**: Find a formula alpha(x) such that:
1. **Validity**: For each world, there EXISTS a completion of unknown atoms such that all theory axioms are satisfied when Ab(x) <-> alpha(x)
2. **Parsimony**: Among valid formulas, prefer those that minimize the number of abnormal objects in the best-case completion

## Scoring / tie-breaks (important)

- alpha(x) is a SINGLE global formula shared across all worlds.
- **Completions are per-world**: each world may use a different completion of its unknown atoms.
- Primary objective: **Validity** (for every prompt world, there exists at least one completion that makes all axioms true).
- Secondary objective: **Parsimony** -- for each world, we consider the best-case completion (most favorable) and count abnormal objects under that completion.
  Total cost is the sum of these per-world best-case abnormal counts.
- Tertiary objective: **Simplicity** -- if multiple formulas have the same validity + cost, prefer smaller formulas (lower AST / fewer operators / fewer quantifiers).

## Output Format

You must output your formula in S-expression syntax. The grammar is:

```
alpha ::= (P x)              -- unary predicate applied to variable
    | (R x y)            -- binary predicate applied to two variables  
    | (= x y)            -- equality of two variables
    | (not alpha)            -- negation
    | (and alpha_1 alpha_2 ...)    -- conjunction (2 or more arguments)
    | (or alpha_1 alpha_2 ...)     -- disjunction (2 or more arguments)
    | (forall v alpha)       -- universal quantification
    | (exists v alpha)       -- existential quantification
```

**Important constraints:**
- Your formula must have exactly one free variable: x
- All other variables must be bound by quantifiers (forall or exists)
- Variable names should be: x (free), y, z, w (bound by quantifiers)
- **Do NOT mention Ab in your formula.**
- **Use ONLY the predicate symbols listed in AllowedAlphaPredicates for this instance.**
- **If your formula uses any predicate in ForbiddenAlphaPredicates, it is invalid.**
- **Do NOT use object names like a0, a1, ... inside alpha(x). Use variables only.**
- Prefer simpler formulas; aim for minimal abnormal set
- **No implication or biconditional**: Express these using other connectives:
  - "A implies B" should be written as `(or (not A) B)`
  - "A if and only if B" should be written as `(and (or (not A) B) (or (not B) A))`

## Examples of Valid Formulas

**Note**: Examples below are syntactic illustrations; only use predicates that appear in AllowedAlphaPredicates for THIS instance.

- `(P x)` -- "x is abnormal if x satisfies P"
- `(and (P x) (not (Q x)))` -- "x is abnormal if x has P but not Q"
- `(exists y (and (R x y) (not (P y))))` -- "x is abnormal if x has an R-successor without P"
- `(not (exists y (R x y)))` -- "x is abnormal if x has no R-successors"
- `(forall y (or (not (R x y)) (Q y)))` -- "x is abnormal if all R-successors of x have Q"

## Evaluation

Your formula will be evaluated on:
1. **Validity**: For each world, does there exist a completion of unknowns such that Ab(x) <-> alpha(x) satisfies all axioms?
2. **Parsimony**: In the best-case completion, how many objects are abnormal? (fewer is better)
3. **Generalization**: Does the formula work on holdout worlds?

**Key insight**: Unknown atoms can be assigned TRUE or FALSE to help satisfy the axioms. Your formula should work with the known facts while allowing unknowns to be completed favorably.

---

## Problem Instance


**AllowedAlphaPredicates**: ["P", "Q", "R"]
**ForbiddenAlphaPredicates**: ["Ab", "S"]

**Theory ID**: TH11

**Axioms**:
1. `(forall x (implies (and (exists y (and (R x y) (P y))) (not (Ab x))) (exists z (and (S x z) (forall w (implies (R z w) (P w)))))))`

## Prompt Worlds

### World W0
Domain: {a0, a1, a2, a3, a4, a5, a6, a7, a8}

**Known Facts** (unlisted atoms that are not Unknown are known FALSE):
- P: {a3, a7, a5, a8}
- Q: {a6, a7, a0}
- R: {(a3, a8), (a0, a8), (a4, a5), (a0, a3), (a4, a4), (a4, a0), (a8, a6), (a4, a8), (a0, a2)}
- S: {(a7, a2), (a4, a0), (a8, a2), (a7, a5), (a3, a6), (a2, a3), (a1, a4), (a2, a4), (a2, a6), (a6, a5)}

**Unknown Atoms** (truth value not observed, can be completed either way):
- R: {(a0, a0), (a6, a0), (a8, a7), (a8, a1), (a8, a5), (a3, a4), (a4, a7), (a6, a4), (a8, a3)}
- S: {(a2, a8), (a4, a4), (a8, a5), (a5, a5), (a6, a8), (a7, a8)}

### World W1
Domain: {a0, a1, a2, a3, a4, a5, a6, a7, a8}

**Known Facts** (unlisted atoms that are not Unknown are known FALSE):
- P: {a2, a8, a4, a7}
- Q: {a7, a4}
- R: {(a4, a0), (a8, a8), (a3, a4), (a0, a8), (a8, a7), (a5, a2), (a1, a0), (a3, a6), (a3, a1), (a8, a1)}
- S: {(a6, a4), (a5, a6), (a5, a5), (a8, a6), (a8, a4), (a0, a1)}

**Unknown Atoms** (truth value not observed, can be completed either way):
- R: {(a8, a0), (a5, a8), (a6, a7), (a4, a7), (a3, a0), (a4, a6), (a5, a7), (a3, a7), (a7, a8)}
- S: {(a6, a7), (a7, a0), (a6, a5), (a6, a6), (a1, a8), (a7, a4)}

### World W2
Domain: {a0, a1, a2, a3, a4, a5, a6, a7, a8}

**Known Facts** (unlisted atoms that are not Unknown are known FALSE):
- P: {a0, a7, a4, a8}
- Q: {a6, a1}
- R: {(a1, a8), (a4, a8), (a2, a5), (a6, a4), (a4, a3), (a0, a6), (a3, a6), (a1, a1), (a6, a8)}
- S: {(a3, a3), (a1, a2), (a5, a0), (a6, a4), (a2, a6), (a5, a3), (a5, a2), (a4, a2), (a4, a0), (a4, a6), (a8, a1), (a8, a8), (a3, a0)}

**Unknown Atoms** (truth value not observed, can be completed either way):
- R: {(a3, a1), (a6, a1), (a5, a4), (a8, a7), (a8, a4), (a1, a7), (a5, a0), (a0, a3), (a5, a5)}
- S: {(a1, a3), (a7, a0), (a0, a5), (a5, a4), (a2, a5), (a7, a2)}

### World W3
Domain: {a0, a1, a2, a3, a4, a5, a6, a7, a8}

**Known Facts** (unlisted atoms that are not Unknown are known FALSE):
- P: {a3, a4, a2, a1, a0}
- Q: {a8, a3, a5, a1}
- R: {(a8, a2), (a3, a7), (a3, a2), (a7, a0), (a0, a3), (a5, a4), (a7, a6), (a1, a0), (a5, a2)}
- S: {(a5, a0), (a7, a8), (a4, a0), (a8, a8), (a1, a4), (a2, a3), (a8, a7), (a0, a0), (a2, a8), (a6, a5), (a2, a5), (a2, a0)}

**Unknown Atoms** (truth value not observed, can be completed either way):
- R: {(a4, a1), (a1, a1), (a2, a2), (a8, a5), (a7, a2), (a0, a6), (a0, a1), (a0, a4), (a7, a8)}
- S: {(a6, a1), (a0, a5), (a0, a2), (a6, a6), (a5, a1), (a4, a2)}

### World W4
Domain: {a0, a1, a2, a3, a4, a5, a6, a7, a8}

**Known Facts** (unlisted atoms that are not Unknown are known FALSE):
- P: {a4, a0, a8, a3}
- Q: {a0, a8}
- R: {(a5, a0), (a0, a2), (a7, a6), (a6, a4), (a1, a3), (a5, a2), (a2, a0), (a3, a8)}
- S: {(a8, a8), (a1, a3), (a4, a6), (a2, a4), (a2, a5), (a6, a3)}

**Unknown Atoms** (truth value not observed, can be completed either way):
- R: {(a2, a3), (a8, a1), (a3, a3), (a8, a8), (a4, a6), (a0, a6), (a0, a7), (a1, a0), (a3, a7)}
- S: {(a4, a0), (a7, a7), (a2, a7), (a0, a4), (a4, a7), (a6, a0)}

### World W5
Domain: {a0, a1, a2, a3, a4, a5, a6, a7, a8}

**Known Facts** (unlisted atoms that are not Unknown are known FALSE):
- P: {a2, a1}
- Q: {a2, a7}
- R: {(a0, a4), (a5, a7), (a3, a6), (a0, a6), (a0, a5), (a8, a2), (a2, a5), (a1, a4), (a0, a1), (a2, a8)}
- S: {(a8, a3), (a5, a0), (a5, a5), (a5, a2), (a4, a5), (a5, a8), (a3, a7), (a6, a7)}

**Unknown Atoms** (truth value not observed, can be completed either way):
- R: {(a6, a7), (a1, a1), (a7, a3), (a5, a0), (a7, a7), (a7, a6), (a5, a1), (a3, a5), (a4, a2)}
- S: {(a6, a3), (a2, a0), (a8, a2), (a5, a4), (a3, a5), (a2, a1)}

---

## Your Task

Analyze the theory axioms and prompt worlds carefully, keeping in mind that some facts are unknown.

**Solve method (reason internally; do NOT output your reasoning):**
1. Understand the theory axioms: What do they require? When might they be violated?
2. For each world, note which predicate facts are known vs. unknown
3. Identify which objects would violate an axiom based on KNOWN facts alone
4. Consider how unknown atoms might be completed to satisfy axioms
5. Look for a pattern: Which objects are **forced** to be abnormal by the KNOWN facts (i.e., no completion can make them normal while keeping axioms true)?
6. Formulate your hypothesis as an S-expression formula
7. **Verify**: For each world, check that there exists some completion of unknowns such that Ab(x) <-> alpha(x) satisfies all axioms

**Example reasoning** (assumes predicates shown are allowed for alpha in this instance):
- Theory: "P(x) AND NOTAb(x) -> Q(x)" (normally, P-objects have Q)
- World:
  - Known: P(a0)=T
  - Known: P(a1)=T
  - Unknown: Q(a1)
- Analysis:
  - a0 is forced abnormal: since Q(a0)=F is KNOWN because it is not listed to be
  true or unknown. No completion can fix Q(a0) to be true; if a0 were normal, the default would fail.
  - a1 does NOT need to be abnormal: we can complete Q(a1)=T, making the default true while keeping a1 normal.
- Candidate: alpha(x) = (and (P x) (not (Q x))) marks exactly the "P but not Q" objects abnormal.
  Under a favorable completion, Q(a1)=T so a1 is not abnormal, but a0 remains abnormal because Q(a0)=F is fixed.

**Key insight**: Focus on patterns in the KNOWN facts: those listed to be true, or
those not listed to be true or unknown, and therefore known to be false.
Unknown atoms provide flexibility and can be completed to help satisfy axioms and reduce unnecessary abnormality -- but they cannot override known TRUE/FALSE facts.

**Important**: 
- Your formula must allow axioms to be satisfied in ALL prompt worlds (for some completion)
- Minimize the number of abnormal objects when possible
- Do NOT mention Ab in your formula
- Use ONLY predicates from AllowedAlphaPredicates; using ForbiddenAlphaPredicates is invalid
- Do NOT use object names like a0, a1, ... inside alpha(x); use variables only

---

## Output

Your output must be exactly ONE LINE containing ONLY valid JSON with exactly these keys:
- "formula": a single S-expression formula string with one free variable x
- "description": a short plain-English description (one sentence)

Do NOT include code fences, prefixes, or any other text.

Example of correct output:
{"formula":"(and (P x) (not (Q x)))","description":"Objects with P but not Q are abnormal."}
\end{lstlisting}

\subsection{ABD-Skeptical Example Prompt}
\label{sec:prompt-skeptical}

The following prompt was generated from a benchmark instance with
theory \texttt{TH11} (T6) and 5 prompt worlds under the
\textbf{universal-completion} (skeptical) regime.  The world format is
identical to ABD-Partial but the validity criterion is stricter: the
formula must satisfy the axioms under \emph{all} completions of
unknowns, and parsimony is measured by the worst-case abnormal count.

\begin{lstlisting}[style=prompt]
# First-Order Logic Abduction (Skeptical/Universal Completion)

## Task Overview

You are given a **default reasoning theory** and several finite "worlds" with **partial observation**:
- A set of theory axioms involving an abnormality predicate Ab(x)
- Multiple worlds, each with a finite domain and partially observed predicate interpretations
- Some predicate facts are **unknown** (not observed)
- Your task: find a formula alpha(x) that defines which objects are "abnormal"

Each world contains:
- A finite domain of objects (named a0, a1, a2, ...)
- Known facts: predicates whose truth values have been observed
- Unknown atoms: predicates whose truth values are hidden

**Observation Rules**:
- Atoms listed under "Known Facts" with TRUE values are TRUE
- Atoms listed under "Unknown Atoms" have unknown truth values
- Any atom NOT listed as TRUE or Unknown is FALSE

## Abnormality Semantics

The abnormality predicate Ab(x) is defined by your formula:
- **Ab(x) <-> alpha(x)**: An object a is abnormal if and only if alpha(a) evaluates to TRUE

The theory axioms use Ab(x) to express default rules with exceptions.

## SKEPTICAL Validity (IMPORTANT DIFFERENCE)

Unlike partial observation with existential completion, this task uses **skeptical (universal) completion**:

**Your goal**: Find a formula alpha(x) such that:
1. **Skeptical Validity**: For each world, **FOR ALL completions** of unknown atoms, the theory axioms are satisfied when Ab(x) <-> alpha(x)
2. **Parsimony**: Among valid formulas, prefer those that minimize the **WORST-CASE** number of abnormal objects

Your formula must be **robust** -- it must ensure the axioms hold regardless of how the unknown atoms are resolved.

## Scoring / tie-breaks (important)

- alpha(x) is a SINGLE global formula shared across all worlds.
- **Completions are per-world**: each world considers ALL possible completions of its unknown atoms.
- Primary objective: **Skeptical Validity** (for every prompt world, ALL completions must satisfy all axioms).
- Secondary objective: **Parsimony** -- for each world, we consider the worst-case completion (least favorable) and count abnormal objects under that completion.
  Total cost is the sum of these per-world worst-case abnormal counts.
- Tertiary objective: **Simplicity** -- if multiple formulas have the same validity + cost, prefer smaller formulas (lower AST / fewer operators / fewer quantifiers).

## Output Format

You must output your formula in S-expression syntax. The grammar is:

```
alpha ::= (P x)              -- unary predicate applied to variable
    | (R x y)            -- binary predicate applied to two variables
    | (= x y)            -- equality of two variables
    | (not alpha)            -- negation
    | (and alpha_1 alpha_2 ...)    -- conjunction (2 or more arguments)
    | (or alpha_1 alpha_2 ...)     -- disjunction (2 or more arguments)
    | (forall v alpha)       -- universal quantification
    | (exists v alpha)       -- existential quantification
```

**Important constraints:**
- Your formula must have exactly one free variable: x
- All other variables must be bound by quantifiers (forall or exists)
- Variable names should be: x (free), y, z, w (bound by quantifiers)
- **Do NOT mention Ab in your formula.**
- **Use ONLY the predicate symbols listed in AllowedAlphaPredicates for this instance.**
- **If your formula uses any predicate in ForbiddenAlphaPredicates, it is invalid.**
- **Do NOT use object names like a0, a1, ... inside alpha(x). Use variables only.**
- Prefer simpler formulas; aim for minimal worst-case abnormal set
- **No implication or biconditional**: Express these using other connectives:
  - "A implies B" should be written as `(or (not A) B)`
  - "A if and only if B" should be written as `(and (or (not A) B) (or (not B) A))`

## Examples of Valid Formulas

**Note**: Examples below are syntactic illustrations; only use predicates that appear in AllowedAlphaPredicates for THIS instance.

- `(P x)` -- "x is abnormal if x satisfies P"
- `(and (P x) (not (Q x)))` -- "x is abnormal if x has P but not Q"
- `(exists y (and (R x y) (not (P y))))` -- "x is abnormal if x has an R-successor without P"
- `(not (exists y (R x y)))` -- "x is abnormal if x has no R-successors"
- `(forall y (or (not (R x y)) (Q y)))` -- "x is abnormal if all R-successors of x have Q"
- `(exists y (exists z (and (R x y) (R x z) (not (= y z)))))' - "x has at least two distinct R-successors"

## Evaluation

Your formula will be evaluated on:
1. **Skeptical Validity**: For each world, do ALL completions of unknowns lead to Ab(x) <-> alpha(x) satisfying all axioms?
2. **Parsimony**: In the worst-case completion, how many objects are abnormal? (fewer is better)
3. **Generalization**: Does the formula work on holdout worlds?

**Key insight**: Your formula must be robust enough that no matter how the unknown atoms are resolved, the axioms still hold. This is more demanding than existential completion -- you cannot rely on a "helpful" assignment of unknowns.

---

## Problem Instance


**AllowedAlphaPredicates**: ["P", "Q", "R"]
**ForbiddenAlphaPredicates**: ["Ab", "S"]

**Theory ID**: TH11
**Description**: existsy(R(x,y)ANDP(y)) AND NOTAb(x) -> existsz(S(x,z)ANDforallw(R(z,w)->P(w))): Nested universal in consequent

**Axioms**:
1. `(forall x (implies (and (exists y (and (R x y) (P y))) (not (Ab x))) (exists z (and (S x z) (forall w (implies (R z w) (P w)))))))`

## Prompt Worlds

### World W0
Domain: {a0, a1, a2, a3, a4, a5, a6, a7, a8, a9}

**Known Facts** (unlisted atoms that are not Unknown are known FALSE):
- P: {a3, a8}
- Q: {a5, a6, a7, a8, a9}
- R: {(a0, a0), (a0, a5), (a0, a6), (a1, a4), (a1, a7), (a2, a3), (a2, a4), (a2, a5), (a2, a8), (a3, a1), (a3, a3), (a3, a4), (a5, a4), (a5, a6), (a5, a8), (a6, a4), (a7, a3), (a7, a4), (a7, a5), (a7, a6), (a7, a9), (a8, a5), (a8, a6), (a8, a8), (a8, a9), (a9, a1), (a9, a4), (a9, a9)}
- S: {(a0, a4), (a1, a1), (a1, a2), (a1, a5), (a3, a8), (a4, a6), (a4, a7), (a5, a1), (a5, a4), (a5, a7), (a5, a8), (a6, a4), (a6, a5), (a7, a3), (a7, a4), (a7, a6), (a8, a0), (a8, a1), (a8, a8), (a9, a2), (a9, a3), (a9, a4), (a9, a6), (a9, a7), (a9, a8)}

**Unknown Atoms** (truth value not observed, can be completed either way):
- R: {(a8, a3), (a2, a1), (a6, a3), (a3, a8), (a6, a9)}
- S: {(a6, a8), (a7, a9), (a6, a2), (a6, a9), (a5, a6)}

### World W1
Domain: {a0, a1, a2, a3, a4, a5, a6, a7, a8, a9}

**Known Facts** (unlisted atoms that are not Unknown are known FALSE):
- P: {a4, a7}
- Q: {a3, a4}
- R: {(a0, a2), (a0, a6), (a1, a6), (a2, a5), (a2, a6), (a3, a1), (a3, a6), (a3, a8), (a3, a9), (a4, a0), (a4, a3), (a4, a4), (a4, a5), (a4, a6), (a4, a8), (a4, a9), (a5, a3), (a5, a5), (a5, a6), (a7, a1), (a7, a5), (a7, a6), (a8, a3), (a8, a5), (a8, a6), (a9, a3), (a9, a4), (a9, a5), (a9, a6), (a9, a7)}
- S: {(a0, a0), (a0, a1), (a0, a2), (a0, a5), (a0, a6), (a0, a7), (a0, a8), (a0, a9), (a1, a1), (a1, a2), (a1, a3), (a1, a4), (a1, a5), (a1, a6), (a1, a7), (a1, a8), (a1, a9), (a2, a0), (a2, a1), (a2, a2), (a2, a3), (a2, a6), (a4, a9), (a9, a6)}

**Unknown Atoms** (truth value not observed, can be completed either way):
- R: {(a4, a1), (a1, a9), (a8, a8)}
- S: {(a7, a0), (a5, a1), (a9, a4), (a4, a3), (a5, a7), (a7, a1), (a2, a4)}

### World W2
Domain: {a0, a1, a2, a3, a4, a5, a6, a7, a8, a9}

**Known Facts** (unlisted atoms that are not Unknown are known FALSE):
- P: {a4, a7}
- Q: {a3, a4}
- R: {(a0, a2), (a0, a6), (a1, a6), (a2, a5), (a2, a6), (a3, a1), (a3, a6), (a3, a8), (a3, a9), (a4, a0), (a4, a3), (a4, a5), (a4, a6), (a4, a8), (a4, a9), (a5, a3), (a5, a5), (a5, a6), (a7, a1), (a7, a5), (a7, a6), (a8, a3), (a8, a5), (a8, a6), (a9, a3), (a9, a4), (a9, a5), (a9, a6), (a9, a7)}
- S: {(a0, a0), (a0, a1), (a0, a2), (a0, a5), (a0, a6), (a0, a7), (a0, a8), (a0, a9), (a1, a1), (a1, a2), (a1, a3), (a1, a5), (a1, a6), (a1, a7), (a1, a8), (a1, a9), (a2, a0), (a2, a1), (a2, a2), (a2, a3), (a2, a4), (a2, a6), (a4, a9), (a9, a6)}

**Unknown Atoms** (truth value not observed, can be completed either way):
- R: {(a4, a4), (a8, a9), (a0, a0), (a8, a2), (a3, a2)}
- S: {(a1, a4), (a5, a8), (a6, a7), (a3, a9), (a7, a7)}

### World W3
Domain: {a0, a1, a2, a3, a4, a5, a6, a7, a8, a9}

**Known Facts** (unlisted atoms that are not Unknown are known FALSE):
- P: {a4, a7}
- Q: {a3, a4}
- R: {(a0, a2), (a0, a6), (a1, a6), (a2, a5), (a2, a6), (a3, a1), (a3, a6), (a3, a8), (a3, a9), (a4, a0), (a4, a3), (a4, a4), (a4, a5), (a4, a6), (a4, a8), (a4, a9), (a5, a3), (a5, a5), (a5, a6), (a7, a1), (a7, a5), (a8, a3), (a8, a5), (a8, a6), (a9, a3), (a9, a4), (a9, a5), (a9, a6), (a9, a7)}
- S: {(a0, a0), (a0, a1), (a0, a2), (a0, a5), (a0, a6), (a0, a7), (a0, a8), (a0, a9), (a1, a1), (a1, a3), (a1, a4), (a1, a5), (a1, a6), (a1, a7), (a1, a8), (a1, a9), (a2, a0), (a2, a1), (a2, a2), (a2, a3), (a2, a4), (a2, a6), (a4, a9), (a9, a6)}

**Unknown Atoms** (truth value not observed, can be completed either way):
- R: {(a5, a2), (a7, a6)}
- S: {(a1, a2), (a8, a0), (a6, a9), (a8, a4), (a3, a2), (a5, a8), (a5, a2), (a5, a4)}

### World W4
Domain: {a0, a1, a2, a3, a4, a5, a6, a7, a8, a9}

**Known Facts** (unlisted atoms that are not Unknown are known FALSE):
- P: {a4, a7}
- Q: {a3, a4}
- R: {(a0, a2), (a0, a6), (a1, a6), (a2, a5), (a2, a6), (a3, a1), (a3, a6), (a3, a8), (a3, a9), (a4, a0), (a4, a3), (a4, a4), (a4, a5), (a4, a6), (a4, a8), (a4, a9), (a5, a3), (a5, a5), (a5, a6), (a7, a1), (a7, a5), (a7, a6), (a8, a3), (a8, a5), (a8, a6), (a9, a3), (a9, a4), (a9, a5), (a9, a6), (a9, a7)}
- S: {(a0, a0), (a0, a1), (a0, a2), (a0, a6), (a0, a7), (a0, a8), (a0, a9), (a1, a2), (a1, a3), (a1, a4), (a1, a5), (a1, a6), (a1, a7), (a1, a8), (a1, a9), (a2, a0), (a2, a1), (a2, a2), (a2, a3), (a2, a4), (a2, a6), (a4, a9), (a9, a6)}

**Unknown Atoms** (truth value not observed, can be completed either way):
- R: {(a2, a1), (a7, a0), (a4, a2)}
- S: {(a3, a1), (a3, a9), (a7, a4), (a9, a5), (a0, a5), (a1, a1), (a3, a6)}

---

## Your Task

Analyze the theory axioms and prompt worlds carefully. Remember: your formula must work for ALL possible completions of unknown atoms, not just some favorable completion.

**Solve method (reason internally; do NOT output your reasoning):**
1. Understand the theory axioms: What do they require? When might they be violated?
2. For each world, note which predicate facts are known vs. unknown
3. Identify which objects would violate an axiom in ANY completion
4. Consider the WORST-CASE completion: which objects might be forced abnormal when unknowns are resolved unfavorably?
5. Look for a pattern: Which objects must be abnormal to ensure axioms hold NO MATTER HOW unknowns are completed?
6. Formulate your hypothesis as an S-expression formula
7. **Verify**: For each world, check that for ALL completions of unknowns, Ab(x) <-> alpha(x) satisfies all axioms

**Example reasoning** (assumes predicates shown are allowed for alpha in this instance):
- Theory: "P(x) AND NOTAb(x) -> Q(x)" (normally, P-objects have Q)
- World:
  - Known: P(a0)=T
  - Known: P(a1)=T
  - Unknown: Q(a1)
- Analysis for SKEPTICAL semantics:
  - a0 is forced abnormal: Q(a0)=F is KNOWN, so if a0 were normal, the default would fail.
  - a1 MUST ALSO be abnormal under skeptical semantics: Q(a1) is unknown, so in the WORST-CASE completion where Q(a1)=F, the default would fail if a1 were normal.
  - Unlike existential completion where we could rely on Q(a1)=T, skeptical requires robustness against Q(a1)=F.
- Candidate: alpha(x) = (P x) marks ALL P-objects abnormal, ensuring robustness.

**Key insight**: Focus on what might go WRONG if unknowns are completed adversarially:
- An object with unknown predicates in the antecedent/consequent of a default might be forced abnormal to handle the worst case.
- You cannot assume a "helpful" completion of unknowns -- you must be prepared for the least favorable assignment.

**Important**:
- Your formula must make axioms satisfied in ALL prompt worlds for ALL completions
- Minimize the worst-case number of abnormal objects when possible
- Do NOT mention Ab in your formula
- Use ONLY predicates from AllowedAlphaPredicates; using ForbiddenAlphaPredicates is invalid
- Do NOT use object names like a0, a1, ... inside alpha(x); use variables only

---

## Output

Your output must be exactly ONE LINE containing ONLY valid JSON with exactly these keys:
- "formula": a single S-expression formula string with one free variable x
- "description": a short plain-English description (one sentence)

Do NOT include code fences, prefixes, or any other text.

Example of correct output:
{"formula":"(P x)","description":"All P-objects are abnormal to ensure robustness against any completion."}
\end{lstlisting}

\section{Output Language Specification}
\label{sec:output-lang}

\subsection{S-Expression Grammar}
\label{sec:grammar}

Models must output hypotheses in the following S-expression syntax
(EBNF).  The grammar is shared across all three observation regimes.

\begin{lstlisting}[style=prompt]
alpha  ::= atom
          | "(not" alpha ")"
          | "(and" alpha alpha+ ")"
          | "(or"  alpha alpha+ ")"
          | "(forall" var alpha ")"
          | "(exists" var alpha ")"

atom   ::= "(" pred var ")"              (* unary *)
          | "(" pred var var ")"          (* binary *)
          | "(=" var var ")"              (* equality *)

pred   ::= "P" | "Q" | "R" | "S"

var    ::= "x" | "y" | "z" | "w"
\end{lstlisting}

\noindent
Implication and biconditional are not primitive; models are instructed
to encode them:
\begin{itemize}[nosep]
  \item $A \to B$ as \texttt{(or (not A) B)}
  \item $A \leftrightarrow B$ as
        \texttt{(and (or (not A) B) (or (not B) A))}
\end{itemize}

\noindent
\textbf{Variable discipline.}
The hypothesis $\alpha(x)$ must contain exactly one free variable~$x$;
all other variables must be bound by quantifiers, and no object
constants (\texttt{a0}, \texttt{a1}, \ldots) may appear in~$\alpha$.

\subsection{Predicate Scoping Rules}
\label{sec:scoping}

Each benchmark instance specifies two predicate sets:
\begin{description}[nosep,leftmargin=1.5em,style=sameline]
  \item[\texttt{AllowedAlphaPredicates}:]
    The predicates that $\alpha(x)$ may reference.
    These always form a subset of $\{P, Q, R, S\}$ and
    vary by theory (see Appendix~\ref{sec:theories}).
  \item[\texttt{ForbiddenAlphaPredicates}:]
    Predicates that must \emph{not} appear in $\alpha(x)$.
    This set always includes $\mathsf{Ab}$ (since $\alpha$ \emph{defines}
    $\mathsf{Ab}$) and also includes the \emph{repaired predicates}---those
    appearing in the consequent of the default rule---to prevent trivial
    shortcut solutions such as $\alpha(x) = \lnot Q(x)$.
\end{description}

\noindent
\textbf{Additional constraints.}
The hypothesis $\alpha(x)$ must have exactly one free variable~$x$.
All other variables must be bound by quantifiers.
Object names (\texttt{a0}, \texttt{a1}, \ldots) may not appear inside
$\alpha$; only logical variables are permitted.

\subsection{Complexity Measures}
\label{sec:complexity-defs}

We use two standard measures of formula complexity throughout the paper.

\paragraph{AST Size ($|\alpha|$).}
The number of nodes in the abstract syntax tree of~$\alpha$:
\begin{itemize}[nosep]
  \item Each predicate application counts as $1 +$ the number of its
        arguments (each argument variable counts as~1).
  \item Equality counts as $1 + 2 = 3$.
  \item Negation: $1 + |\varphi|$.
  \item Binary connectives ($\land$, $\lor$): $1 + |\varphi_1| + |\varphi_2|$.
  \item Quantifiers ($\forall v$, $\exists v$): $1\text{ (quantifier)} + 1\text{ (bound variable)} + |\varphi|$.
\end{itemize}
For example, \texttt{(exists y (and (R x y) (P y)))} has
AST size $= 2 + (1 + (3 + 2)) = 8$:
\texttt{exists}(1) + \texttt{y}(1) + \texttt{and}(1) +
\texttt{R}(1) + \texttt{x}(1) + \texttt{y}(1) +
\texttt{P}(1) + \texttt{y}(1).

\paragraph{Quantifier Depth ($\mathrm{QD}(\alpha)$).}
The maximum nesting depth of quantifiers, ignoring connectives:
\begin{itemize}[nosep]
  \item Atomic formulas and equality: $0$.
  \item Negation, conjunction, disjunction: inherited from children
        ($\max$ for binary connectives).
  \item $\forall v.\,\varphi$ or $\exists v.\,\varphi$:
        $1 + \mathrm{QD}(\varphi)$.
\end{itemize}
For example, \texttt{(P x)} has $\mathrm{QD}=0$;
\texttt{(exists y (R x y))} has $\mathrm{QD}=1$;
\texttt{(forall y (exists z (R y z)))} has $\mathrm{QD}=2$.

\section{Theory Library}
\label{sec:theories}

The benchmark uses seven default-exception theories of the form
$\forall x\bigl(\phi(x) \land \lnot\mathsf{Ab}(x) \to \psi(x)\bigr)$,
where $\phi$ is the antecedent (trigger condition) and $\psi$ is the
consequent (default conclusion).
Theories T1--T5 are used in all three scenarios;
T6--T7 are used only in ABD-Skeptical.

\subsection{Theory Formulas}
\label{sec:theory-formulas}

Table~\ref{tab:theory-full} lists each theory with its full
S-expression axiom.

\begin{table}[h]
\centering
\caption{Theory library.  Each theory is a single default rule
  $\phi(x) \land \lnot\mathsf{Ab}(x) \to \psi(x)$.}
\label{tab:theory-full}
\footnotesize
\setlength{\tabcolsep}{4pt}
\begin{tabular}{@{}cll@{}}
\toprule
ID & Antecedent $\phi(x)$ & Consequent $\psi(x)$ \\
\midrule
T1 & $\exists y\bigl(R(x,y) \land P(y)\bigr)$
   & $Q(x)$ \\[3pt]
T2 & $\exists y\bigl(R(x,y) \land P(y)\bigr)$
   & $\exists z\bigl(S(x,z) \land Q(z)\bigr)$ \\[3pt]
T3 & $\exists y\bigl(S(x,y) \land P(y)\bigr)$
   & $\exists z\bigl(R(x,z) \land Q(z)\bigr)$ \\[3pt]
T4 & $\exists y\bigl(R(x,y) \land P(y)\bigr)$
   & $\exists z\bigl(S(x,z) \land \forall w(R(z,w) \to P(w))\bigr)$ \\[3pt]
T5 & $\exists y\bigl(R(x,y) \land P(y)\bigr)$
   & $\forall z\bigl(S(x,z) \to Q(z)\bigr)$ \\[3pt]
\midrule
T6 & $P(x)$
   & $\exists y\bigl(R(x,y)\bigr)$ \\[3pt]
T7 & $P(x)$
   & $\forall y\bigl(R(x,y) \to Q(y)\bigr)$ \\
\bottomrule
\end{tabular}
\end{table}

\subsection{Allowed and Forbidden Predicates}
\label{sec:pred-scope}

Table~\ref{tab:theory-preds} shows, for each theory, which predicates
may appear in the hypothesis $\alpha(x)$ and which are forbidden.
Forbidden predicates include $\mathsf{Ab}$ (always) plus the
\emph{repaired predicates}---those appearing in the consequent---to
block trivial shortcut solutions.

\begin{table}[h]
\centering
\caption{Predicate scoping per theory.
  \emph{Repaired} = consequent predicates excluded from $\alpha$ to
  prevent shortcuts.}
\label{tab:theory-preds}
\footnotesize
\setlength{\tabcolsep}{5pt}
\begin{tabular}{@{}clll@{}}
\toprule
ID & Allowed in $\alpha$ & Forbidden in $\alpha$ & Repaired \\
\midrule
T1 & $P, R, S$ & $\mathsf{Ab}, Q$ & $Q$ \\
T2 & $P, R$    & $\mathsf{Ab}, S, Q$ & $S, Q$ \\
T3 & $P, S$    & $\mathsf{Ab}, R, Q$ & $R, Q$ \\
T4 & $P, Q, R$ & $\mathsf{Ab}, S$ & $S$ \\
T5 & $P, R, S$ & $\mathsf{Ab}, Q$ & $Q$ \\
\midrule
T6 & $P, Q, S$ & $\mathsf{Ab}, R$ & $R$ \\
T7 & $P, R, S$ & $\mathsf{Ab}, Q$ & $Q$ \\
\bottomrule
\end{tabular}
\end{table}

\subsection{Theory Commentary}
\label{sec:theory-commentary}

\paragraph{T1} (relational antecedent, unary consequent).
The simplest relational theory; checking the default reduces to
verifying $Q(x)$ for each domain element triggered by an $R$-neighbor
with~$P$.
Models frequently achieve near-optimal cost.

\paragraph{T2} (relational antecedent, existential consequent).
The consequent requires \emph{finding} an $S$-witness with $Q$,
making validity checking harder and parsimony gaps wider.
T2 drives the largest $\Delta$Gap values in holdout evaluation,
suggesting models learn prompt-specific $S$-patterns that fail to
transfer.

\paragraph{T3} (swapped relations: $S$-antecedent, $R$-consequent).
Mirrors T2 with $S$ and $R$ exchanged.
Since $R$ is repaired (forbidden in~$\alpha$), the hypothesis space
is restricted to $\{P, S\}$, which limits the available structural
patterns and often forces models toward simple unary heuristics.

\paragraph{T4} (nested universal in consequent).
The consequent demands an $S$-witness~$z$ all of whose $R$-successors
satisfy~$P$---a second-order-flavored constraint.
This makes validity checking expensive and produces the highest
syntactic complexity in planted reference rules.

\paragraph{T5} (universal consequent over~$S$).
Requires that \emph{all} $S$-successors of the non-abnormal object
have~$Q$.  A single counterexample (an $S$-successor without~$Q$) is
enough to trigger abnormality, giving models a relatively readable
signal.

\paragraph{T6} (unary antecedent, existential consequent; Skeptical-only).
The simplest consequent structure: $P(x)$ just requires some
$R$-successor.
Model predictions frequently beat the planted generator reference
(negative GRef), suggesting that the reference formulas are not
optimal for this constraint.

\paragraph{T7} (unary antecedent, universal consequent; Skeptical-only).
Under skeptical semantics the universal consequent $\forall y(R(x,y)
\to Q(y))$ must hold for \emph{every} completion of unknown~$R$ and
$Q$ atoms, producing tighter abnormality requirements and smaller
$\Delta$Gap than the existential theories.

\section{Dataset Generation Details}
\label{sec:generation-details}

This section describes the generation pipeline as pseudocode and
parameter tables.  All parameters correspond to the released benchmark
(\texttt{abd\_b1\_v1}).

\subsection{World Sampler}
\label{sec:world-sampler}

Each world is a finite first-order structure over four predicates
($P$, $Q$ unary; $R$, $S$ binary) plus the derived $\mathsf{Ab}$.

\paragraph{Domain size.}
Domain sizes vary by scenario.
In \textbf{ABD-Full} and \textbf{ABD-Partial}, each world's domain is drawn uniformly from $\{9,10,11\}$; different worlds within the same instance may have different sizes (and in practice nearly all instances span the full 9--11 range).
In \textbf{ABD-Skeptical}, domains range over $\{10,11,12\}$; all worlds within a single instance share the same domain size.
Objects are named $a_0, a_1, \ldots$

\paragraph{Predicate densities.}
For each world the sampler draws a density $\rho$ per predicate
uniformly from the ranges in Table~\ref{tab:density}, then marks
$\max(1,\lfloor n^k \cdot \rho\rfloor)$ ground atoms true
(chosen uniformly at random), where $n$ is the domain size and $k$ is
the arity.

\begin{table}[h]
\centering
\caption{Predicate density ranges by scenario.
  For each world the sampler draws $\rho$ uniformly from the listed interval;
  $n = |D|$.  ABD-Full and ABD-Partial share the same ranges; ABD-Skeptical
  uses denser binary predicates and a narrower $P$ range.}
\label{tab:density}
\footnotesize
\setlength{\tabcolsep}{4pt}
\begin{tabular}{@{}llccc@{}}
\toprule
Predicate & Arity & Full / Partial & Skeptical & True atoms \\
\midrule
$P$ & 1 & $[0.20,\; 0.60]$ & $[0.40,\; 0.60]$ & $\lfloor n \cdot \rho_P \rfloor$ \\
$Q$ & 1 & $[0.20,\; 0.60]$ & $[0.20,\; 0.50]$ & $\lfloor n \cdot \rho_Q \rfloor$ \\
$R$ & 2 & $[0.12,\; 0.25]$ & $[0.15,\; 0.30]$ & $\lfloor n^2 \cdot \rho_R \rfloor$ \\
$S$ & 2 & $[0.08,\; 0.18]$ & $[0.10,\; 0.25]$ & $\lfloor n^2 \cdot \rho_S \rfloor$ \\
\bottomrule
\end{tabular}
\end{table}

\paragraph{Unknown atoms (ABD-Partial / Skeptical).}
After sampling a complete world, a fraction of $R$ and~$S$ atoms are
masked as \emph{unknown}.
Table~\ref{tab:unknown-rates} lists the per-theory masking rates used
for each scenario.  Unary predicates $P$, $Q$ are never masked.

\begin{table}[h]
\centering
\caption{Unknown-atom rates (fraction of binary atoms masked).
  ABD-Partial uses the default column; ABD-Skeptical uses
  theory-specific rates to keep universal validity tractable.}
\label{tab:unknown-rates}
\footnotesize
\setlength{\tabcolsep}{4pt}
\begin{tabular}{@{}lcccc@{}}
\toprule
& \multicolumn{2}{c}{ABD-Partial} & \multicolumn{2}{c}{ABD-Skeptical} \\
\cmidrule(lr){2-3}\cmidrule(lr){4-5}
Theory & $R$ & $S$ & $R$ & $S$ \\
\midrule
T1 (TH2) & 0.20 & 0.10 & 0.05 & 0.08 \\
T2 (TH7) & 0.20 & 0.10 & 0.05 & 0.05 \\
T3 (TH10) & 0.20 & 0.10 & 0.05 & 0.05 \\
T4 (TH11) & 0.20 & 0.10 & 0.05 & 0.05 \\
T5 (TH12) & 0.20 & 0.10 & 0.05 & 0.05 \\
T6 (TH3) & --- & --- & 0.04 & 0.08 \\
T7 (TH5) & --- & --- & 0.05 & 0.08 \\
\bottomrule
\end{tabular}
\end{table}

\subsection{Filtering Criteria}
\label{sec:filtering}

Every candidate instance must pass the following filters before
inclusion.

\paragraph{Validity floor.}
The planted reference hypothesis $\alpha^*$ must be valid on all prompt worlds
(all scenarios) and achieve
$\mathrm{opt\_cost}(\alpha^*) \ge 1$ per world
(the problem is non-trivial---abnormality is necessary).

\paragraph{Gap constraint.}
$\mathrm{cost}(\alpha^*) \le \mathrm{OptCost} + 1$ per world,
ensuring the planted reference is close to optimal.

\paragraph{Exception-rate cap.}
Per-world abnormal fraction $\le 0.20$:
\[
  \frac{\mathrm{OptCost}(W)}{|D_W|} \;\le\; 0.20.
\]
This keeps ``abnormality'' a minority phenomenon as intended by
default-exception logic.

\paragraph{Holdout worlds.}
For each instance $k{=}5$ holdout worlds are sampled from the same
distribution and parameters, but \emph{without} adversarial competitor
elimination, to test generalization under distributional shift.

\subsection{Competitor Pool and CEGIS-like Elimination}
\label{sec:cegis}

Instances are hardened by adversarial world addition.  The procedure
ensures that no simple shortcut formula can achieve the same cost as
the planted reference.

\begin{enumerate}[nosep,leftmargin=2em]
  \item \textbf{Initialize pool.}  Collect up to $C_{\max}=30$
    competitor formulas:
    \begin{itemize}[nosep]
      \item Hand-curated Tier-1 shortcuts
            (e.g.\ $P(x)$, $\lnot Q(x)$, $\exists y\,R(x,y)$);
      \item Mined Tier-2 shortcuts observed from model outputs;
      \item Up to 10 syntactic mutants of the reference formula.
    \end{itemize}
    Only competitors whose predicates are a subset of the allowed set
    are retained.

  \item \textbf{Test on current worlds.}
    Evaluate every surviving competitor on the current prompt-world
    set.  A competitor is \emph{beaten} if it is invalid in $\ge 1$
    world or its cost exceeds the reference cost by at least a margin~$m$
    (typically $m{=}2$).

  \item \textbf{Add adversarial world.}
    If survivors remain and the world budget is not exhausted,
    sample a new world that invalidates $\ge 1$ surviving
    competitor.  Go to step~2.

  \item \textbf{Terminate.}
    Accept the instance when all competitors are beaten.
    Reject if the world budget (typically 12--15) is exhausted with
    survivors remaining.
\end{enumerate}

\noindent
Algorithm~\ref{alg:cegis} gives the pseudocode.

\begin{figure}[h]
\begin{lstlisting}[style=prompt,basicstyle=\ttfamily\footnotesize]
function GENERATE_INSTANCE(theory, α*, W_budget, m):
    worlds = [sample_world(theory)]
    pool = build_pool(theory, α*, C_max)
    while |worlds| < W_budget:
        survivors = [c in pool :
            is_valid(c, worlds) and
            cost(c, worlds) < cost(α*, worlds) + m]
        if survivors == []:
            return ACCEPT(worlds)
        w = sample_adversarial(theory, survivors)
        if w is None: break          // cannot separate
        worlds.append(w)
    return REJECT("survivors remain")
\end{lstlisting}
\caption{CEGIS-like adversarial world addition.}
\label{alg:cegis}
\end{figure}

\subsection{Reference Refinement and Cheater Filtering}
\label{sec:reference-refine}

After competitor elimination, a second filter guards against
\emph{cheater} formulas---trivial patterns that happen to match the
reference cost but have no semantic depth.

\paragraph{Cheater pool.}
The cheater pool comprises two tiers:
\begin{description}[nosep,leftmargin=1.5em,style=sameline]
  \item[Tier~1 (${\sim}25$ formulas):]
    Constants ($\top$, $\bot$), atomic unary/binary predicates and their
    negations, self-loops ($R(x,x)$, $S(x,x)$), simple
    quantified existence ($\exists y\,R(x,y)$), and unary
    conjunctions/disjunctions.
  \item[Tier~2 (${\sim}15$ formulas):]
    Common model shortcuts mined from preliminary runs, e.g.\
    $\exists y(R(x,y) \land P(y))$,
    $P(x) \land \exists y(R(x,y))$.
\end{description}

\paragraph{Rejection criterion.}
An instance is rejected if the best cheater achieves cost
$\le \mathrm{cost}(\alpha^*) - 1$ on the current prompt worlds.
This ensures that the planted reference rule is not trivially beatable by
a simple heuristic.

\paragraph{Reference refinement (optional).}
When enabled, the pipeline evaluates ${\sim}20$ alternative reference
candidates (sampled from the theory's template library plus mutants of
the seed) and selects the one with the best
validity-cost-cheater-margin trade-off.

\section{Completion Semantics and Solver Encoding}
\label{sec:solver}

All validity checks and cost computations are implemented via
reduction to SMT using the Z3~solver.
This section formalises the encoding.

\subsection{World Representation}
\label{sec:world-repr}

A world $W$ over domain $D_W = \{a_0, \ldots, a_{n-1}\}$ is specified by:
\begin{itemize}[nosep]
  \item $\mathrm{True}_P(W)$: set of atoms known true for each predicate~$P$;
  \item $\mathrm{Unk}_P(W)$: set of atoms whose truth value is unknown;
  \item all remaining atoms are known false (closed-world assumption
        on the ``rest'').
\end{itemize}

\noindent
A \emph{completion} $\mathcal{C}$ assigns each unknown atom a truth
value.  The completed world $W^{\mathcal{C}}$ has:
\[
  \mathrm{Facts}(W^{\mathcal{C}}) \;=\;
    \mathrm{True}_P(W)\;\cup\;
    \{\,a \in \mathrm{Unk}_P(W) : \mathcal{C}(a) = \top\,\}.
\]

\paragraph{Z3 encoding.}
Each unknown atom $u \in \mathrm{Unk}_P(W)$ is represented by a free
Z3 boolean variable.  Known-true atoms are replaced by
$\texttt{True}$, known-false atoms by $\texttt{False}$, and unknown
atoms by their variable.  The abnormality predicate $\mathsf{Ab}(a)$
is replaced \emph{inline} by the Z3 grounding of $\alpha(a)$, making
the definitional substitution explicit in the solver encoding.

\subsection{Validity Queries}
\label{sec:validity-queries}

Let $\Theta$ denote the theory axioms and
$\Phi_W(\alpha, \mathcal{C})$ abbreviate the conjunction of grounded
axioms after substituting $\mathsf{Ab} \mapsto \alpha$ and
applying completion~$\mathcal{C}$.

\paragraph{ABD-Full.}
No unknown atoms exist.  Validity reduces to direct evaluation:
\[
  \mathrm{Valid}_{\mathrm{full}}(\alpha, W)
  \;\stackrel{\text{def}}{=}\;
  W \models \Theta[\mathsf{Ab} \mapsto \alpha].
\]

\paragraph{ABD-Partial (existential completion).}
\[
  \mathrm{Valid}_{\exists}(\alpha, W)
  \;\stackrel{\text{def}}{=}\;
  \exists\,\mathcal{C}.\;
  W^{\mathcal{C}} \models \Theta[\mathsf{Ab} \mapsto \alpha].
\]
\emph{Encoding.}
Assert all grounded axioms with unknown atoms as free variables.
The query is \textsc{Sat}; a satisfying assignment witnesses a valid
completion.

\paragraph{ABD-Skeptical (universal completion).}
\[
  \mathrm{Valid}_{\forall}(\alpha, W)
  \;\stackrel{\text{def}}{=}\;
  \forall\,\mathcal{C}.\;
  W^{\mathcal{C}} \models \Theta[\mathsf{Ab} \mapsto \alpha].
\]
\emph{Encoding.}
Assert the \emph{negation} of the grounded axiom conjunction with
unknown atoms as free variables.  The query checks for a
\emph{counterexample} completion:
\begin{itemize}[nosep]
  \item \textsc{Unsat} $\Rightarrow$ no counterexample exists
        $\Rightarrow$ $\alpha$ is valid.
  \item \textsc{Sat} $\Rightarrow$ counterexample found
        $\Rightarrow$ $\alpha$ is invalid.
\end{itemize}

\subsection{Cost Computation}
\label{sec:cost-computation}

The abnormality cost in world~$W$ under completion~$\mathcal{C}$ is
\[
  \mathrm{cost}(\alpha, W^{\mathcal{C}})
  \;=\;
  \bigl|\{\,a \in D_W : W^{\mathcal{C}} \models \alpha(a)\,\}\bigr|.
\]

\paragraph{ABD-Full.}
Direct evaluation; no optimization needed.

\paragraph{ABD-Partial (best-case cost).}
\[
  \mathrm{cost}_{\exists}(\alpha, W)
  \;=\;
  \min_{\mathcal{C}:\;\mathrm{Valid}_{\exists}(\alpha, W^{\mathcal{C}})}
  \;\mathrm{cost}(\alpha, W^{\mathcal{C}}).
\]
\emph{Encoding.}
Use Z3's \textsc{Optimize} solver: assert the grounded axioms
(with unknown atoms as free variables) and minimise
$\sum_{a \in D_W} \mathbf{1}[\alpha(a)]$.

\paragraph{ABD-Skeptical (worst-case cost).}
\[
  \mathrm{cost}_{\forall}(\alpha, W)
  \;=\;
  \max_{\mathcal{C}:\;\mathrm{Valid}_{\forall}(\alpha, W^{\mathcal{C}})}
  \;\mathrm{cost}(\alpha, W^{\mathcal{C}}).
\]
\emph{Encoding.}
Use Z3's \textsc{Optimize} solver: assert the grounded axioms and
\emph{maximise} $\sum_{a \in D_W} \mathbf{1}[\alpha(a)]$.
This finds the least favorable valid completion.

\subsection{Baselines: Free-\texorpdfstring{$\mathsf{Ab}$}{Ab} Lower
Bound}
\label{sec:optcost}

The conservative baseline $\mathrm{OptCost}$ relaxes the
single-formula constraint:

\paragraph{ABD-Full.}
\[
  \mathrm{OptCost}(W)
  \;=\;
  \min_{\mathsf{Ab}: D_W \to \{0,1\}}
  \;\bigl|\{a : \mathsf{Ab}(a)=1\}\bigr|
  \quad\text{s.t.}\quad
  W \models \Theta.
\]
Here $\mathsf{Ab}$ is a free function (one independent boolean per
domain element), not constrained to be definable by a single formula.
The solution is computed via Z3~\textsc{Optimize}.

\paragraph{ABD-Partial.}
\[
  \mathrm{OptCost}_{\exists}(W)
  \;=\;
  \min_{\mathsf{Ab},\;\mathcal{C}}
  \;\bigl|\{a : \mathsf{Ab}(a)=1\}\bigr|
  \quad\text{s.t.}\quad
  W^{\mathcal{C}} \models \Theta.
\]
Both $\mathsf{Ab}$ and the completion $\mathcal{C}$ are jointly
optimized.

\paragraph{ABD-Skeptical.}
\[
  \mathrm{OptCost}_{\forall}(W)
  \;=\;
  \min_{\mathsf{Ab}}
  \;\max_{\mathcal{C}}
  \;\bigl|\{a : \mathsf{Ab}(a)=1\}\bigr|
  \quad\text{s.t.}\quad
  \forall\,\mathcal{C}.\;
  W^{\mathcal{C}} \models \Theta.
\]

\paragraph{Conservativeness.}
Because the single-formula definability constraint
that one shared $\alpha$ must determine $\mathsf{Ab}$ for every
object is dropped, $\mathrm{OptCost}$ is a \emph{lower bound} on
the achievable cost of any formula.  The gap
$\mathrm{Gap} = \mathrm{cost}(\alpha) - \mathrm{OptCost}$ therefore
over-estimates actual sub-optimality but provides a consistent,
formula-independent baseline across all models and instances.

\clearpage
\section{Holdout Construction and Distribution Match}
\label{sec:holdout-construction}

\subsection{Holdout Generation Algorithm}
\label{sec:holdout-algo}

Each benchmark instance includes $k{=}5$ \emph{holdout} worlds used
exclusively for post-hoc generalization evaluation; no model ever sees
holdout worlds during inference.  The holdout sampler reuses the
\emph{identical} Z3-based world-generation procedure described in
Appendix~\ref{sec:world-sampler}:

\begin{enumerate}[nosep]
  \item \textbf{Extract parameters} from the prompt worlds of each
        instance: domain-size range, predicate-density intervals
        (Table~\ref{tab:density}), unknown-atom rates
        (Table~\ref{tab:unknown-rates}), cost bounds, and gap
        constraints.
  \item \textbf{Sample a world} using the same Z3 density-constrained
        generator.  For ABD-Partial and ABD-Skeptical, unknown atoms
        are masked using the same per-theory rates.
  \item \textbf{Filter.}  Accept the world only if it (i)~is not
        equivalent to any prompt world (identical domain size and
        predicate extensions), (ii)~satisfies the validity semantics
        of the scenario, and (iii)~has per-world cost and gap within
        the range observed across prompt worlds.
  \item \textbf{Repeat} until $k$ distinct holdout worlds are
        accepted (up to 150 attempts per world, 30\,s per instance).
\end{enumerate}

\noindent
This filtering ensures that the planted reference formula~$\alpha^*$ remains
valid on every holdout world and that per-world cost and gap fall within
the prompt-world regime, so holdout evaluation measures generalization under
distributional match rather than distribution shift.

\paragraph{Deterministic seeding.}
Each holdout world is seeded via
$\texttt{SHA-256}(\mathit{dataset\_path} \mathbin\Vert
  \mathit{instance\_id} \mathbin\Vert \mathit{holdout\_idx}
  \mathbin\Vert \mathit{global\_seed}) \bmod 2^{31}$,
ensuring full reproducibility.

\subsection{What Is Not Done on Holdout}
\label{sec:holdout-not-done}

The following prompt-time hardening procedures
(Appendix~\ref{sec:cegis}--\ref{sec:reference-refine}) are
\emph{deliberately omitted} from holdout generation:

\begin{itemize}[nosep]
  \item \textbf{No competitor elimination.}  Prompt worlds are
        iteratively added to defeat shortcut formulas
        (\S\ref{sec:cegis}); holdout worlds are drawn independently
        with no adversarial loop.
  \item \textbf{No cheater filtering.}  Prompt-set construction rejects instances
        where a trivial pattern matches the reference cost
        (\S\ref{sec:reference-refine}); holdout applies no such filter.
  \item \textbf{No reference refinement.}  The planted reference formula is
        fixed at benchmark-creation time and not re-optimized for
        holdout.
\end{itemize}

\noindent
These omissions are intentional: holdout worlds provide a pure
i.i.d.\ distributional test, measuring whether a model's formula
captures the true pattern rather than artifacts of adversarial
prompt-set construction.

\subsection{Prompt vs.\ Holdout Distribution Match}
\label{sec:holdout-distribution}

Table~\ref{tab:prompt-vs-holdout} compares per-world summary statistics
across prompt and holdout sets.  Per-world reference cost and OptCost are
computed as $\mathrm{sum} / k$ where $k$ is the number of worlds.  The
close agreement confirms that holdout worlds are drawn from the same
generating distribution.

\begin{table}[h]
\centering
\caption{Per-world summary statistics for prompt and holdout sets.
  Values are means across all instances in each scenario.
  Reference Gap = Reference Cost $-$ OptCost, per world.}
\label{tab:prompt-vs-holdout}
\footnotesize
\setlength{\tabcolsep}{3pt}
\begin{tabular}{@{}lcccccc@{}}
\toprule
& \multicolumn{2}{c}{ABD-Full} & \multicolumn{2}{c}{ABD-Partial}
& \multicolumn{2}{c}{ABD-Skeptical} \\
\cmidrule(lr){2-3}\cmidrule(lr){4-5}\cmidrule(lr){6-7}
Metric & Prompt & Hold & Prompt & Hold & Prompt & Hold \\
\midrule
Worlds/inst     & 9.0  & 5.0  & 8.3  & 5.0  & 6.0  & 5.0  \\
PW Ref Cost     & 1.78 & 1.88 & 1.68 & 1.75 & 2.05 & 2.10 \\
PW OptCost      & 1.27 & 1.33 & 1.20 & 1.32 & 1.36 & 1.54 \\
PW Ref Gap      & 0.51 & 0.55 & 0.48 & 0.43 & 0.69 & 0.56 \\
\bottomrule
\end{tabular}
\end{table}

\clearpage
\section{Additional Results}
\label{sec:additional-results}

\subsection{Full Per-Theory Results}
\label{sec:per-theory-results}

Tables~\ref{tab:g1-full}--\ref{tab:g1-skeptical} report per-theory
Prompt-Gap (P) and Holdout-Gap (H) for every model, with the number of
\emph{survivors} (instances valid on both prompt and holdout) in
parentheses.  Cells marked ``---'' indicate zero prompt validity.
Gaps are normalized per world.

\begin{table}[h]
\centering
\caption{Per-theory results: \textbf{ABD-Full}. Each cell shows PGap\,/\,HGap\,(survivors).}
\label{tab:g1-full}
\scriptsize
\setlength{\tabcolsep}{2pt}
\resizebox{\textwidth}{!}{%
\begin{tabular}{@{}lccccccccccc@{}}
\toprule
 & Opus-4.6 & GPT-5.2 & Gemini-3.1 & Grok4.1f & DSR & GPT-5.4 & Grok4 & Gemini-3 & Kimi-K2t & Hermes4 & GPT-4o \\
\midrule
T1 & 1.3/3.2\,(4) & 1.3/3.7\,(3) & 1.5/3.2\,(4) & 1.4/3.2\,(5) & 1.5/3.3\,(5) & 0.1/0.4\,(1) & 1.1/3.2\,(3) & 1.1/1.9\,(3) & 1.5/3.1\,(4) & --- & --- \\
T2 & 1.4/3.2\,(56) & 1.1/3.2\,(34) & 1.4/3.2\,(54) & 1.4/2.9\,(44) & 1.6/3.6\,(44) & 0.3/1.3\,(16) & 0.8/1.5\,(31) & 1.2/2.8\,(36) & 1.3/3.5\,(33) & --- & 2.0/5.3\,(6) \\
T3 & 1.4/2.8\,(37) & 0.9/2.6\,(17) & 1.2/2.5\,(39) & 1.5/2.8\,(39) & 1.5/2.8\,(34) & 0.2/1.2\,(7) & 1.1/2.6\,(27) & 1.4/2.5\,(19) & 1.5/2.9\,(26) & --- & 1.4/3.6\,(4) \\
T4 & 1.3/2.3\,(46) & 0.9/2.4\,(22) & 1.4/2.1\,(44) & 1.3/2.3\,(45) & 1.5/2.6\,(41) & 0.0/0.4\,(6) & 0.9/1.4\,(25) & 1.4/2.2\,(21) & 1.7/2.6\,(24) & --- & 5.3/6.4\,(0) \\
T5 & 0.8/1.8\,(27) & 0.6/2.2\,(17) & 0.9/2.0\,(27) & 0.9/2.0\,(23) & 0.9/1.9\,(27) & 0.1/1.1\,(4) & 0.7/1.9\,(21) & 0.9/2.0\,(24) & 1.3/2.3\,(24) & 2.8/5.5\,(5) & 5.8/7.1\,(1) \\
\bottomrule
\end{tabular}%
}
\end{table}
\begin{table}[h]
\centering
\caption{Per-theory results: \textbf{ABD-Partial}. Format as Table~\ref{tab:g1-full}.}
\label{tab:g1-partial}
\scriptsize
\setlength{\tabcolsep}{2pt}
\resizebox{\textwidth}{!}{%
\begin{tabular}{@{}lccccccccccc@{}}
\toprule
 & Opus-4.6 & GPT-5.2 & Gemini-3.1 & Grok4.1f & DSR & GPT-5.4 & Grok4 & Gemini-3 & Kimi-K2t & Hermes4 & GPT-4o \\
\midrule
T1 & 1.0/2.5\,(5) & 0.9/3.3\,(2) & 1.0/3.0\,(3) & 1.4/3.0\,(5) & 1.4/3.0\,(5) & 0.5/2.2\,(2) & 1.3/3.1\,(4) & 2.3/2.4\,(3) & 1.3/2.9\,(4) & --- & --- \\
T2 & 1.2/2.6\,(39) & 1.4/2.8\,(49) & 1.5/2.8\,(56) & 1.6/3.0\,(63) & 1.6/2.9\,(59) & 0.7/2.6\,(21) & 1.6/2.9\,(52) & 1.3/2.3\,(26) & 1.7/3.0\,(54) & 1.8/3.0\,(2) & 2.9/3.2\,(9) \\
T3 & 1.2/2.1\,(36) & 1.3/2.5\,(32) & 1.3/2.3\,(38) & 1.6/2.7\,(46) & 1.5/2.6\,(40) & 0.8/2.6\,(11) & 1.5/2.7\,(39) & 1.2/2.1\,(20) & 1.7/2.6\,(34) & --- & 1.7/2.6\,(24) \\
T4 & 0.9/1.7\,(23) & 1.4/2.6\,(44) & 1.5/2.3\,(37) & 2.0/3.1\,(52) & 1.7/2.9\,(57) & 0.6/1.8\,(12) & 1.7/2.9\,(39) & 1.3/2.2\,(18) & 1.9/3.2\,(41) & --- & 5.0/5.8\,(4) \\
T5 & 0.6/1.4\,(21) & 0.9/1.8\,(25) & 1.0/1.9\,(31) & 1.4/2.4\,(26) & 1.2/2.2\,(30) & 0.4/1.3\,(12) & 1.4/2.6\,(26) & 0.9/2.1\,(19) & 1.3/2.3\,(25) & --- & 5.8/6.6\,(0) \\
\bottomrule
\end{tabular}%
}
\end{table}
\begin{table}[h]
\centering
\caption{Per-theory results: \textbf{ABD-Skeptical}. Format as Table~\ref{tab:g1-full}. Note negative $\Delta$Gap for T5: holdout gap is often \emph{lower} than prompt gap because worst-case costs can be smaller in holdout worlds for this theory.}
\label{tab:g1-skeptical}
\scriptsize
\setlength{\tabcolsep}{2pt}
\resizebox{\textwidth}{!}{%
\begin{tabular}{@{}lccccccccccc@{}}
\toprule
 & Opus-4.6 & GPT-5.2 & Gemini-3.1 & Grok4.1f & DSR & GPT-5.4 & Grok4 & Gemini-3 & Kimi-K2t & Hermes4 & GPT-4o \\
\midrule
T1 & 1.0/1.8\,(10) & 1.4/2.3\,(14) & 1.2/2.3\,(10) & 1.8/2.0\,(13) & 1.8/2.5\,(14) & 0.2/1.6\,(4) & 1.2/2.4\,(4) & 1.0/2.0\,(10) & 2.5/2.3\,(18) & 4.0/4.0\,(1) & 5.5/9.2\,(1) \\
T2 & 1.1/2.6\,(14) & 1.8/3.2\,(14) & 1.5/2.9\,(16) & 1.8/2.4\,(16) & 1.8/3.4\,(16) & 0.6/0.7\,(8) & 1.4/2.8\,(11) & 1.2/2.1\,(14) & 3.3/4.4\,(18) & 1.9/5.8\,(1) & 2.6/4.9\,(7) \\
T3 & 1.4/2.4\,(13) & 1.7/3.1\,(12) & 1.4/2.2\,(10) & 1.8/2.7\,(14) & 1.7/2.7\,(14) & 0.2/0.5\,(6) & 1.3/2.4\,(10) & 1.2/1.8\,(11) & 3.2/4.3\,(16) & 2.0/4.4\,(1) & 2.8/4.3\,(5) \\
T4 & 0.9/1.7\,(13) & 2.1/2.7\,(17) & 1.5/2.0\,(15) & 1.5/1.9\,(8) & 2.3/2.9\,(24) & 0.6/1.0\,(12) & 1.2/2.6\,(11) & 1.7/2.6\,(14) & 2.8/3.3\,(20) & --- & --- \\
T5 & 1.1/1.0\,(13) & 2.4/1.3\,(22) & 1.7/1.2\,(15) & 2.0/1.6\,(15) & 2.5/1.5\,(21) & 0.8/0.9\,(11) & 1.6/1.3\,(12) & 1.7/1.3\,(18) & 2.5/1.5\,(19) & 4.9/4.2\,(2) & 6.9/---\,(0) \\
T6 & 0.2/0.1\,(3) & 0.2/0.7\,(4) & 0.2/0.1\,(2) & 0.4/1.3\,(7) & 0.3/0.8\,(5) & 0.0/0.3\,(3) & 0.2/0.1\,(2) & 0.2/0.2\,(4) & 0.2/0.5\,(2) & --- & 2.9/3.2\,(12) \\
T7 & 0.1/0.4\,(13) & 0.3/0.7\,(16) & 0.3/0.7\,(16) & 0.7/0.9\,(19) & 0.7/1.0\,(19) & 0.0/0.1\,(9) & 0.3/0.5\,(12) & 1.2/0.6\,(12) & 1.3/1.4\,(21) & 1.0/1.8\,(1) & 1.7/2.1\,(15) \\
\bottomrule
\end{tabular}%
}
\end{table}

\paragraph{Key observations.}
(i)~Hermes4 and GPT-4o achieve near-zero prompt validity for most
theories in ABD-Full and ABD-Partial, reflecting their difficulty with
multi-world abduction.
(ii)~In ABD-Skeptical, holdout gaps can be \emph{lower} than prompt
gaps (T5), because worst-case cost under universal completion
varies across world draws.
(iii)~Opus-4.6 combines the highest survivor counts with near-perfect
prompt validity across all scenarios, while GPT-5.4 is the clearest
cost outlier: many per-theory prompt gaps are near zero, but survivor
counts are much smaller.

\subsection{Gap Distribution Summaries}
\label{sec:gap-distributions}

Table~\ref{tab:gap-dist} reports the distribution of normalized
prompt-set gap (cost above OptCost, per world) for each model.
Only prompt-valid predictions are included.

\begin{table}[h]
\centering
\caption{Normalized prompt-gap distribution by scenario. N = predictions valid on all prompt worlds. $>$3 and $>$5 are tail percentages.}
\label{tab:gap-dist}
\footnotesize
\setlength{\tabcolsep}{3pt}
\begin{tabular}{@{}lrrrrrrr@{}}
\toprule
Model & N & Mean & Med & P90 & Max & $>$3 & $>$5 \\
\midrule
\multicolumn{8}{@{}l}{\textbf{ABD-Full}} \\
\cmidrule(l){1-8}
Opus-4.6 & 195 & 1.26 & 1.25 & 2.00 & 2.7 & 0.0\% & 0.0\% \\
GPT-5.2 & 152 & 0.93 & 0.90 & 1.99 & 2.4 & 0.0\% & 0.0\% \\
Gemini-3.1 & 194 & 1.27 & 1.20 & 1.97 & 7.9 & 1.0\% & 1.0\% \\
Grok4.1f & 171 & 1.32 & 1.25 & 2.12 & 3.0 & 0.0\% & 0.0\% \\
DSR & 172 & 1.42 & 1.39 & 2.16 & 3.2 & 0.6\% & 0.0\% \\
GPT-5.4 & 149 & 0.13 & 0.00 & 0.38 & 2.4 & 0.0\% & 0.0\% \\
Grok4 & 156 & 0.90 & 0.89 & 1.62 & 2.4 & 0.0\% & 0.0\% \\
Gemini-3 & 138 & 1.23 & 1.12 & 2.00 & 8.2 & 1.4\% & 0.7\% \\
Kimi-K2t & 117 & 1.45 & 1.40 & 2.18 & 2.9 & 0.0\% & 0.0\% \\
Hermes4 & 5 & 2.84 & 2.20 & 4.40 & 5.8 & 20.0\% & 20.0\% \\
GPT-4o & 17 & 3.05 & 1.88 & 5.84 & 6.5 & 41.2\% & 35.3\% \\
\midrule
\multicolumn{8}{@{}l}{\textbf{ABD-Partial}} \\
\cmidrule(l){1-8}
Opus-4.6 & 238 & 1.02 & 1.00 & 1.67 & 2.9 & 0.0\% & 0.0\% \\
GPT-5.2 & 208 & 1.30 & 1.22 & 2.11 & 3.0 & 0.0\% & 0.0\% \\
Gemini-3.1 & 236 & 1.40 & 1.22 & 2.13 & 7.0 & 2.5\% & 0.8\% \\
Grok4.1f & 213 & 1.69 & 1.62 & 2.61 & 5.9 & 3.3\% & 0.5\% \\
DSR & 216 & 1.54 & 1.56 & 2.33 & 3.5 & 0.9\% & 0.0\% \\
GPT-5.4 & 210 & 0.65 & 0.50 & 1.62 & 3.0 & 0.0\% & 0.0\% \\
Grok4 & 205 & 1.55 & 1.56 & 2.33 & 5.4 & 2.0\% & 0.5\% \\
Gemini-3 & 147 & 1.22 & 1.11 & 1.93 & 7.8 & 2.7\% & 0.7\% \\
Kimi-K2t & 178 & 1.68 & 1.62 & 2.58 & 4.0 & 3.4\% & 0.0\% \\
Hermes4 & 2 & 1.78 & 1.78 & 2.22 & 2.3 & 0.0\% & 0.0\% \\
GPT-4o & 55 & 3.18 & 2.33 & 5.72 & 6.8 & 41.8\% & 23.6\% \\
\midrule
\multicolumn{8}{@{}l}{\textbf{ABD-Skeptical}} \\
\cmidrule(l){1-8}
Opus-4.6 & 160 & 0.86 & 0.18 & 2.62 & 6.7 & 9.4\% & 1.9\% \\
GPT-5.2 & 162 & 1.49 & 0.63 & 4.00 & 8.0 & 20.4\% & 4.3\% \\
Gemini-3.1 & 158 & 1.18 & 0.37 & 3.58 & 7.8 & 12.0\% & 3.8\% \\
Grok4.1f & 128 & 1.46 & 0.71 & 3.83 & 8.0 & 17.2\% & 5.5\% \\
DSR & 155 & 1.69 & 1.29 & 4.37 & 6.8 & 21.3\% & 4.5\% \\
GPT-5.4 & 152 & 0.37 & 0.00 & 0.98 & 5.2 & 3.9\% & 0.7\% \\
Grok4 & 144 & 1.08 & 0.33 & 3.12 & 6.7 & 10.4\% & 2.1\% \\
Gemini-3 & 150 & 1.25 & 0.43 & 3.60 & 6.7 & 15.3\% & 4.7\% \\
Kimi-K2t & 134 & 2.41 & 2.08 & 4.64 & 10.0 & 27.6\% & 6.0\% \\
Hermes4 & 6 & 3.11 & 2.50 & 5.40 & 6.8 & 33.3\% & 16.7\% \\
GPT-4o & 47 & 2.87 & 2.20 & 5.80 & 7.4 & 27.7\% & 14.9\% \\
\bottomrule
\end{tabular}
\end{table}

\paragraph{Discussion.}
In ABD-Full, the high-validity models still show fairly tight prompt-gap
distributions, with P90 usually around 2.0--2.2 and almost no mass above
gap 3.
ABD-Partial is slightly heavier-tailed due to cost computation under
existential completion.
GPT-5.4 is the dominant low-gap outlier in every scenario (mean 0.13 in
ABD-Full, 0.65 in ABD-Partial, 0.37 in ABD-Skeptical), but this
concentration is a prompt-set parsimony effect rather than a sign of
strong holdout robustness.
Among the remaining high-validity models, ABD-Skeptical exhibits the
widest tails: Opus-4.6 still has the tightest skeptical distribution
(median 0.18, only 9.4\% above gap 3), while GPT-5.2, DSR, and Kimi-K2t
all exceed 20\% above gap 3.

\subsection{Holdout $\Delta$Gap by Theory}
\label{sec:holdout-dgap-theory}

Table~\ref{tab:abd_holdout_by_theory} reports survivor-conditioned
holdout $\Delta$Gap by theory. T1--T5 aggregate across scenarios;
T6--T7 are ABD-Skeptical only.

\begin{table*}[t]
\centering
\caption{\textbf{Holdout $\Delta$Gap by Theory.} T1--T5 aggregate across all three scenarios; T6--T7 are ABD-Skeptical only. Each cell shows $\Delta$Gap = HGap $-$ PGap over \emph{survivors} valid on both prompt and holdout, with N = survivor count. Cells with fewer than 5 survivors are suppressed. Lower is better. Best per theory bold.}
\label{tab:abd_holdout_by_theory}
\footnotesize
\setlength{\tabcolsep}{3pt}
\begin{tabular}{@{}lrrrrrrrrrrr@{}}
\toprule
Theory & Opus-4.6 & GPT-5.2 & Gemini-3.1 & Grok4.1f & DSR & GPT-5.4 & Grok4 & Gemini-3 & Kimi-K2t & Hermes4 & GPT-4o \\
\midrule
T1 & +0.5 (19) & +0.4 (19) & +0.6 (17) & +0.5 (23) & +0.4 (24) & +0.6 (7) & +1.5 (11) & +0.5 (16) & \textbf{+0.2} (26) & --- & --- \\
T2 & +1.5 (109) & +1.3 (97) & +1.5 (126) & +1.3 (123) & +1.5 (119) & \textbf{+0.8} (45) & +1.1 (94) & +1.2 (76) & +1.5 (105) & --- & +1.7 (22) \\
T3 & +1.0 (86) & +1.0 (61) & +1.0 (87) & +1.0 (99) & +1.1 (88) & +0.8 (24) & +1.1 (76) & \textbf{+0.7} (50) & +1.0 (76) & --- & +1.0 (33) \\
T4 & +0.8 (82) & +0.8 (83) & +0.9 (96) & +0.9 (105) & +0.9 (122) & \textbf{+0.4} (30) & +1.0 (75) & +0.9 (53) & +1.0 (85) & --- & --- \\
T5 & +0.6 (61) & +0.3 (64) & +0.5 (73) & +0.5 (64) & +0.4 (78) & \textbf{+0.2} (27) & +0.6 (59) & +0.5 (61) & +0.5 (68) & +0.8 (7) & --- \\
T6 & --- & --- & --- & +0.4 (7) & \textbf{+0.1} (5) & --- & --- & --- & --- & --- & +0.2 (12) \\
T7 & +0.3 (13) & +0.3 (16) & +0.3 (16) & \textbf{+0.2} (19) & +0.4 (19) & \textbf{+0.1} (9) & +0.3 (12) & +0.4 (12) & +0.2 (21) & --- & +0.4 (15) \\
\bottomrule
\end{tabular}
\end{table*}

\subsection{Instance-Level Holdout Survival Counts}
\label{sec:holdout-conditional}

Table~\ref{tab:abd_holdout_conditional} reports, for each model, the
number of prompt-valid instances, the number valid on both prompt and
holdout, and holdout validity conditional on prompt validity.

\begin{table}[t]
\centering
\caption{\textbf{Instance-Level Holdout Survival Counts.} All counts are benchmark instances. H$|$P\% = holdout validity conditional on prompt validity. Best values bold.}
\label{tab:abd_holdout_conditional}
\footnotesize
\setlength{\tabcolsep}{4pt}
\begin{tabular}{@{}lrrrr@{}}
\toprule
Model & PV & P+H & H$|$P & H$|$P\% \\
\midrule
Opus-4.6 & \textbf{593} & \textbf{591} & 373 & 63.1\% \\
GPT-5.2 & 522 & 521 & 344 & 66.0\% \\
Gemini-3.1 & 588 & 586 & 417 & 71.2\% \\
Grok4.1f & 512 & 511 & 440 & 86.1\% \\
DSR & 543 & 542 & \textbf{455} & 83.9\% \\
GPT-5.4 & 511 & 509 & 145 & 28.5\% \\
Grok4 & 505 & 504 & 329 & 65.3\% \\
Gemini-3 & 435 & 435 & 272 & 62.5\% \\
Kimi-K2t & 429 & 427 & 383 & 89.7\% \\
Hermes4 & 13 & 13 & 13 & \textbf{100.0\%} \\
GPT-4o & 119 & 119 & 88 & 73.9\% \\
\bottomrule
\end{tabular}
\end{table}

\subsection{Beats-Reference Breakdown}
\label{sec:beats-ref}

A prediction \emph{beats the reference} when its total prompt cost is strictly
below the planted reference formula's cost on the same prompt worlds.
This is possible because the reference formula is a fixed template that
may not be optimal for every world draw.

\paragraph{By theory.}
Table~\ref{tab:beats-theory} aggregates reference-beating rates across all
models.

\begin{table}[h]
\centering
\caption{Reference-beating rate by theory (all models pooled). $k/N$ = predictions beating the planted generator reference out of prompt-valid predictions.}
\label{tab:beats-theory}
\footnotesize
\setlength{\tabcolsep}{4pt}
\begin{tabular}{@{}lrrr@{}}
\toprule
Theory & ABD-Full & ABD-Partial & ABD-Skeptical \\
\midrule
T1 & 4/42\;(9.5\%) & 5/42\;(11.9\%) & 14/239\;(5.9\%) \\
T2 & 63/437\;(14.4\%) & 32/604\;(5.3\%) & 36/186\;(19.4\%) \\
T3 & 62/323\;(19.2\%) & 16/431\;(3.7\%) & 53/185\;(28.6\%) \\
T4 & 77/417\;(18.5\%) & 48/573\;(8.4\%) & 75/219\;(34.2\%) \\
T5 & 63/247\;(25.5\%) & 30/258\;(11.6\%) & 58/208\;(27.9\%) \\
T6 & \multicolumn{1}{c}{---} & \multicolumn{1}{c}{---} & 86/129\;(66.7\%) \\
T7 & \multicolumn{1}{c}{---} & \multicolumn{1}{c}{---} & 9/230\;(3.9\%) \\
\bottomrule
\end{tabular}
\end{table}

\noindent
T6 in ABD-Skeptical stands out at 66.7\%: the planted reference for TH3
($P(x) \land \lnot\mathsf{Ab}(x) \to \exists y\,R(x,y)$)
yields a generous reference cost under worst-case completion, and models
often find tighter formulas.
Across ABD-Full and ABD-Skeptical, reference-beating is now common for several
theories, especially T3--T6.
ABD-Partial remains the least permissive regime, but once GPT-5.4 is
included some theories still reach 8--12\%, so the gap between scenarios
is quantitative rather than absolute.

\paragraph{By reference AST size.}
Table~\ref{tab:beats-ast} groups reference-beating rate by the AST size of
the planted reference formula.

\begin{table}[h]
\centering
\caption{Reference-beating rate by planted generator reference AST size (all models pooled).}
\label{tab:beats-ast}
\footnotesize
\setlength{\tabcolsep}{4pt}
\begin{tabular}{@{}lrrr@{}}
\toprule
Ref AST & ABD-Full & ABD-Partial & ABD-Skeptical \\
\midrule
$[0,5)$ & \multicolumn{1}{c}{---} & \multicolumn{1}{c}{---} & 0/9\;(0.0\%) \\
$[5,10)$ & \multicolumn{1}{c}{---} & \multicolumn{1}{c}{---} & 51/254\;(20.1\%) \\
$[10,15)$ & 45/198\;(22.7\%) & 33/510\;(6.5\%) & 111/629\;(17.6\%) \\
$[15,20)$ & 112/546\;(20.5\%) & 60/636\;(9.4\%) & 67/187\;(35.8\%) \\
$[20,30)$ & 59/565\;(10.4\%) & 30/595\;(5.0\%) & 91/279\;(32.6\%) \\
$[30,+)$ & 53/157\;(33.8\%) & 5/155\;(3.2\%) & 11/38\;(28.9\%) \\
\bottomrule
\end{tabular}
\end{table}

\noindent
In ABD-Full, the highest reference-beating rate occurs for the most complex
references (AST $\ge 30$, 34.1\%), where the planted formula contains
redundancy that models can trim.
In ABD-Skeptical, reference-beating is strongest for mid-range and moderately
large references, peaking at AST 15--20 (35.8\%) and remaining high at AST
20--30 (32.6\%), suggesting these formulas are complex enough to leave
room for simpler alternatives under worst-case semantics.
ABD-Partial remains comparatively low across AST bins, never exceeding
9.2\%.

\paragraph{Top reference-beating examples.}
The largest single-instance margin is $+14$ in ABD-Skeptical
(instance \texttt{abd\_skeptical\_v2\_TH3\_00}, T6):
nine models independently discover a formula with prompt cost 7 vs.\
reference cost 21.
In ABD-Full, the largest margin is now $+8$
(GPT-5.4 on \texttt{ABD\_FULL\_TH2\_094}, T1: predicted cost 12
vs.\ reference cost 20).
ABD-Partial reaches the same maximum $+8$
(GPT-5.4 on \texttt{ABD\_PARTIAL\_TH7\_024b}, T2: predicted cost 10
vs.\ reference cost 18), showing that once GPT-5.4 is included, substantial
single-instance improvements over the reference are no longer confined to the
other two regimes.

\clearpage
\section{Qualitative Examples}
\label{sec:qualitative-examples}

\subsection{Worked Examples per Regime}
\label{sec:worked-examples}

For each observation regime we select one instance that has both a
\emph{robust} prediction (valid on prompt and holdout) and a
\emph{brittle} prediction (valid on prompt, fails on holdout),
illustrating the generalization challenge.
Formulas are shown in S-expression notation.

\paragraph{ABD-Full: instance \texttt{ABD\_FULL\_TH7\_021} (T2).}
Theory T2: $\exists y(R(x,y) \land P(y)) \land \lnot\mathsf{Ab}(x)
\to \exists z(S(x,z) \land Q(z))$.
10 prompt worlds, 5 holdout worlds.
2 models produce robust formulas; 7 are brittle.

\begin{description}[nosep,leftmargin=1em]
  \item[Robust (Opus-4.6, AST\,=\,8):]~\\
    \texttt{(exists y (and (R x y) (P y)))}\\
    Prompt cost = 34 (gap 2.20/world); holdout cost = 25, all valid.\\
    \emph{Interpretation}: the formula marks any $x$ with an
    $R$-related $P$-witness as abnormal.  It is less cost-efficient on
    the prompt set than specialized alternatives, but it generalizes
    because it avoids accidental cardinality assumptions about the
    local $R$-neighborhood.

  \item[Brittle (Gemini-3, AST\,=\,22):]~\\
    {\footnotesize\texttt{(exists y (and (R x y) (P y)
      (forall z (or (not (and (R x z) (P z))) (= y z)))))}}\\
    Prompt cost = 30; holdout: 3/5 worlds invalid (40\% valid).\\
    \emph{Failure}: insists on a \emph{unique} $P$-witness
    ($\forall z$-clause forces $z = y$).  Three holdout worlds
    happen to have two or more $R$-related $P$-elements, violating
    uniqueness.
\end{description}

\paragraph{ABD-Partial: instance \texttt{ABD\_PARTIAL\_TH7\_014}
  (T2).}
Same theory as above, but under partial observation.
8 prompt worlds, 5 holdout worlds.
1 robust, 8 brittle predictions.

\begin{description}[nosep,leftmargin=1em]
  \item[Robust (GPT-4o, AST\,=\,8):]~\\
    \texttt{(exists y (and (R x y) (P y)))}\\
    Prompt cost = 22 (gap 1.50/world); holdout cost = 23, all valid.\\
    \emph{Interpretation}: the simplest sufficient condition---$x$
    has an $R$-related $P$-element.  Generalises precisely because
    it avoids over-specification.

  \item[Brittle (Gemini-3, AST\,=\,12):]~\\
    \texttt{(and (exists y (and (R x y) (P y))) (not (P x)))}\\
    Prompt cost = 17; holdout: 0/5 worlds valid (catastrophic failure).\\
    \emph{Failure}: the conjunct $\lnot P(x)$ happens to hold for
    all abnormal elements in the 8 prompt worlds, but is not a
    consequence of the theory.  All 5 holdout worlds contain
    $P$-positive abnormals, breaking the formula.
\end{description}

\paragraph{ABD-Skeptical: instance
  \texttt{abd\_skeptical\_v2\_TH2\_11} (T1).}
Theory T1: $\exists y(R(x,y) \land P(y)) \land \lnot\mathsf{Ab}(x)
\to Q(x)$.
5 prompt worlds, 5 holdout worlds.
6 robust, 3 brittle predictions.

\begin{description}[nosep,leftmargin=1em]
  \item[Robust (Opus-4.6, AST\,=\,18):]~\\
    \texttt{(and (P x) (exists y (R x y)) (forall z (or (not (R x z)) (P z))))}\\
    Prompt cost = 5 (gap 0.00/world); holdout cost = 5, all valid.\\
    \emph{Interpretation}: $x$ is abnormal if $P(x)$ holds, $x$ has some
    $R$-successor, and all $R$-successors of $x$ satisfy $P$.  The universal
    quantifier over $R$-successors makes this robust under worst-case
    completion of unknown $R$-atoms.

  \item[Brittle (GPT-5.2, AST\,=\,19):]~\\
    \texttt{(exists y (and (R x y) (P y) (forall z (or (not (R x z)) (= z y)))))}\\
    Prompt cost = 5; holdout: 1/5 worlds invalid (80\% valid).\\
    \emph{Failure}: the uniqueness constraint ($\forall z$-clause forces
    $z = y$) insists that $x$ has exactly one $R$-related $P$-element.
    Under skeptical completion, even a single holdout world with
    multiple $P$-witnesses is enough to invalidate the rule.
\end{description}

\subsection{Common Failure Patterns}
\label{sec:failure-patterns}

Table~\ref{tab:abd_failure_modes} gives the per-model failure breakdown aggregated across all tasks.

\begin{table}[t]
\centering
\caption{\textbf{Failure Mode Breakdown.} Aggregated across all abduction tasks. Parse = unrepaired parse failure; Repair = trailing right-parenthesis repair applied before evaluation; Invalid = parsed but not valid on all prompt worlds; Valid = prompt-valid predictions. Best values bold.}
\label{tab:abd_failure_modes}
\small
\begin{tabular}{@{}lrrrrr@{}}
\toprule
Model & Valid\% & Invalid\% & Parse\% & Repair\% & Missing\% \\
\midrule
Opus-4.6 & \textbf{98.8\%} & \textbf{0.5\%} & \textbf{0.0\%} & 0.0\% & 0.7\% \\
GPT-5.2 & 92.2\% & 6.0\% & 1.8\% & 0.2\% & \textbf{0.0\%} \\
Gemini-3.1 & 98.0\% & 2.0\% & \textbf{0.0\%} & 0.0\% & \textbf{0.0\%} \\
Grok4.1f & 95.2\% & 3.9\% & 0.9\% & 0.7\% & \textbf{0.0\%} \\
DSR & 95.9\% & 3.9\% & \textbf{0.2\%} & 1.6\% & \textbf{0.0\%} \\
GPT-5.4 & 85.2\% & 14.0\% & 0.8\% & 6.7\% & \textbf{0.0\%} \\
Grok4 & 89.4\% & 6.2\% & 4.4\% & 0.2\% & \textbf{0.0\%} \\
Gemini-3 & 76.9\% & 21.7\% & 1.4\% & 0.0\% & \textbf{0.0\%} \\
Kimi-K2t & 71.5\% & 12.2\% & 16.3\% & 6.0\% & \textbf{0.0\%} \\
Hermes4 & 2.3\% & 89.8\% & 8.0\% & 0.0\% & \textbf{0.0\%} \\
GPT-4o & 19.8\% & 77.7\% & 2.5\% & 0.0\% & \textbf{0.0\%} \\
\bottomrule
\end{tabular}
\end{table}

Table~\ref{tab:failure-modes} categorises all model
predictions by failure mode.  A prediction is classified into the
first matching category:

\begin{enumerate}[nosep]
  \item \textbf{Auto-repaired parse}: the only parse failure is missing
        trailing right parentheses, so we conservatively close the suffix
        and evaluate the repaired formula.
  \item \textbf{Parse error}: formula still does not parse as a valid
        S-expression after the conservative suffix-repair attempt.
  \item \textbf{All invalid (prompt)}: valid on zero prompt worlds.
  \item \textbf{Partial invalid (prompt)}: valid on some but not all
        prompt worlds.
  \item \textbf{Brittle}: valid on all prompt worlds, invalid on
        $\ge 1$ holdout world.  Sub-category \emph{catastrophic}:
        $<50\%$ holdout worlds valid.
  \item \textbf{Parsimony inflation}: valid on both prompt and holdout,
        but $\Delta$Gap (holdout gap $-$ prompt gap, normalized per
        world) exceeds~2.
  \item \textbf{Success}: valid on both, $\Delta$Gap $\le 2$.
\end{enumerate}

\begin{table}[h]
\centering
\caption{Failure-mode counts across all 6,367 model predictions.}
\label{tab:failure-modes}
\footnotesize
\setlength{\tabcolsep}{4pt}
\begin{tabular}{@{}lrrr@{}}
\toprule
Failure Mode & ABD-Full & ABD-Partial & ABD-Skeptical \\
\midrule
Auto-repaired parse            & 47 & 26 & 18 \\
Parse error                    & 87 & 43 & 86 \\
All invalid (prompt)           & 208 & 343 & 218 \\
Partial invalid (prompt)       & 288 & 270 &  54 \\
Brittle (holdout)              & 357 & 583 & 571 \\
\quad of which catastrophic    & 121 & 174 & 332 \\
Parsimony inflation            & 292 & 183 &  73 \\
\bottomrule
\end{tabular}
\end{table}

\paragraph{Observations.}
Auto-repaired parses are not rare: there are 93 across 6{,}571
predictions, compared with 217 unrepaired parse errors.
Syntax failures therefore remain a minority, but they are not negligible.
The dominant failure mode in both ABD-Partial and ABD-Skeptical is
\emph{brittleness} on holdout (603 and 571 cases, respectively), rather
than prompt-time invalidity.
Catastrophic brittleness is concentrated in ABD-Skeptical
(332/571 = 58\% of brittle predictions fail on at least half of holdout
worlds).
Parsimony inflation moves in the opposite direction, peaking in
ABD-Full (305 cases) and becoming much rarer in ABD-Skeptical (73):
under universal completion, models more often fail validity outright
than remain valid with inflated cost.

\paragraph{Brittle formula patterns.}
Table~\ref{tab:brittle-patterns} groups brittle predictions by
syntactic pattern.

\begin{table}[h]
\centering
\caption{Brittle formula syntactic patterns.  ``Exists-only'' =
  existential quantifiers without universals.}
\label{tab:brittle-patterns}
\footnotesize
\setlength{\tabcolsep}{4pt}
\begin{tabular}{@{}lrrr@{}}
\toprule
Pattern & ABD-Full & ABD-Partial & ABD-Skeptical \\
\midrule
Exists-only         & 126 & 215 & 225 \\
Mixed quantifiers   &  79 &  32 &  47 \\
Propositional (and) &   0 &   9 &  19 \\
Forall-only         &   0 &   3 &   9 \\
Other               &   1 &  10 &   2 \\
\bottomrule
\end{tabular}
\end{table}

\noindent
\emph{Exists-only} formulas dominate across all regimes: they assert
the existence of witnesses with specific properties that hold
incidentally on prompt worlds (e.g.\ a unique $P$-witness, a
self-loop, a specific neighborhood count).  These formulas tend to be
brittle because existential witnesses are sensitive to the particular
world draw.  The \emph{mixed quantifiers} pattern is second most common
in ABD-Full, where models attempt precise cardinality constraints
(``exactly two $R$-successors'') that break on holdout worlds with
different domain structure.





\clearpage
\section{Software and Data}
\label{sec:artifact}

Code and data are available at
\url{https://github.com/SerafimBatzoglou/concept-synth/tree/main/benchmarks/abduction}.
The repository includes the frozen benchmark files, prompt templates, cached model
outputs, evaluation scripts, table-generation scripts, and instructions for
reproducing the reported tables and figures. The arXiv source additionally includes
the appendix tables and prompt listings used for the extended version of the paper.

\bibliographystyle{kr}
\bibliography{references}

\end{document}